\pdfoutput=1

\documentclass[11pt]{article}

\usepackage[final]{acl}

\usepackage{times}
\usepackage{latexsym}

\usepackage[T1]{fontenc}

\usepackage[utf8]{inputenc}

\usepackage{microtype}

\usepackage{inconsolata}

\usepackage{graphicx}
\usepackage{graphicx}
\usepackage{microtype}
\usepackage{graphicx}
\usepackage{subfigure}
\usepackage{booktabs} 
\usepackage{tcolorbox}
\usepackage{listings}
\usepackage{fancyvrb}
\usepackage{lipsum}
\usepackage{amsmath}
\usepackage{siunitx}
 \usepackage{multirow}
\usepackage{tabularx}   
\usepackage{booktabs}
\usepackage{multirow}
\usepackage{array}
\usepackage{booktabs}
\usepackage{array}
\usepackage{makecell}
\usepackage{algorithm}
\usepackage{algpseudocode}
\usepackage{amsmath} 
\usepackage{amssymb}
\usepackage[table]{xcolor}

\tcbuselibrary{breakable, listings}
\usepackage{listings}

\usepackage{graphicx}
\usepackage{enumitem}
\title{
    \textsc{CogTRL}: Training LLMs for Scientific Discovery Assistance using \\ Cognitive Traces via Reinforcement Learning
}

\author{Shrinidhi Kumbhar \quad Santosh Mashetty \quad  Divij Handa \quad Kevin Coutinho \\  \textbf{Siddharth Sambhaji Ghule} \quad  \textbf{Chitta Baral} \\
  Arizona State University\\
\small{\texttt{\{skumbha4, chitta\}@asu.edu}}
}

\begin{document}
\maketitle
\begin{abstract}
Large Language Models (LLMs) trained on extensive scientific research are increasingly integrated as assistants for scientific discovery. However, most research papers omit the fine-grained cognitive process of examining constraints, failed alternatives, and iterative decisions required to achieve the desired goal. Such cognitive processes are vital for real-world scientists working toward specific goals under constraints. In this paper, we show that LLMs, when trained to produce such cognitive traces, perform better as scientific discovery assistants than when trained solely on scientific literature. We propose \textsc{CogTRL}, a trajectory-level reinforcement learning framework that trains LLMs to emulate cognitively grounded reasoning by jointly optimizing cognitive traces and the scientific steps produced in an interleaved manner. Across two 3B-parameter models and two scientific domains (AI and Materials Science), \textsc{CogTRL} improves method quality by an average of 7.85 points over comparable 3B model baselines and achieves competitive performance relative to 70B parameter models. Moreover, analysis by domain experts shows a preference for methods generated by \textsc{CogTRL} over the baselines.
\end{abstract}

\begingroup
\renewcommand\thefootnote{}
\footnotetext{\url{https://github.com/shri071/CoGTRL}}
\addtocounter{footnote}{-1}
\endgroup

\begin{quote}
    \textit{Science is a way of thinking much more than it is a body of facts.}
    \hfill- Carl Sagan
\end{quote}

\begin{figure*}[t]
    \centering
    \includegraphics[width=\textwidth]{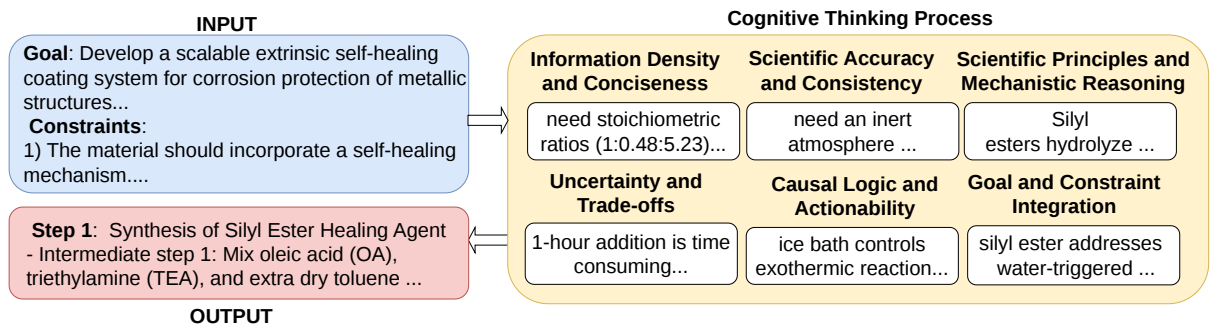}
\caption{
Given a research goal and constraints, the LLM simulates a cognitive thought process and generates a step-by-step methodology by producing cognitive traces before each method step. Step $N$ is conditioned on traces and steps from stages $1$ to $N\!-\!1$, and generation continues until the goal is achieved and all constraints are satisfied.
}
    \label{fig:Teaser}
\end{figure*}

\section{Introduction}

Large Language Models (LLMs) are widely used as assistants in several domains, such as coding \citep{novikov2025alphaevolve}, planning \citep{parmar2025plangen}, and computer-use \citep{awadallah2025fara}. One such domain where LLMs show promise to accelerate research is scientific discovery \citep{ai4science2023impact, gottweis2025towards, krolik2024towards,yamada2025ai,yang2024moose}. For a scientific discovery assistance task, given a goal, for example, \textit{``Develop a self-healing coating for offshore environments''} and constraints, for example, \textit{``Should be eco-friendly''}, the model is tasked with generating a step-by-step methodology for achieving the goal while satisfying the constraints \citep{ai4science2023impact, kumbhar2025hypothesis}. The generated method can aid in guiding a human researcher in achieving the given goal and constraints. Such assistance has the potential to drastically reduce the time and labor costs of methodological design.

While recent research has shown promise of using LLMs in scientific discovery \citep{ai4science2023impact, gottweis2025towards,sprueill2024chemreasoner, wang2023scientific, merchant2023scaling}, they are primarily focused on agentic approaches using frontier LLMs applied to narrow scientific tasks such as information extraction, property prediction, literature synthesis, and hypotheses generation \citep{yang2024moose, xiong2024improving, tong2024automating, ghafarollahi2024sciagents, sprueill2024chemreasoner, ding2024matexpert, shojaee2024llm, yang2023large}. In contrast, we focus on training open-source LLMs for the scientific discovery assistance task. While LLMs are traditionally trained on a wide variety of scientific data, including the latest research articles and papers, these data sources often omit the intermediate cognitive processes underlying scientific decision-making. Since scientific discovery, by nature, is a fundamentally cognitive process, in which researchers hypothesize, evaluate constraints, and iteratively refine their plans \citep{nersessian2025scientists, sep_scientific_discovery, thagard2012cognitive, williams2010role, klahr2000exploring}, instilling such behaviors while training is crucial.

In this paper, we propose a method to train LLMs for scientific discovery assistance with Cognitive Traces via Reinforcement Learning (\textsc{CogTRL}). \textsc{CogTRL} treats the task as a trajectory-level problem and uses Group Relative Policy Optimization (GRPO) \citep{shao2024deepseekmath} as its reinforcement learning backbone. Given a research goal and a set of constraints, the model generates multiple rollouts, each consisting of a sequence of cognitive traces and corresponding methodological steps conditioned on those traces. The structure of \textsc{CogTRL} trajectories is illustrated in Figure~ \ref{fig:Teaser}. Each cognitive trace explicitly articulates why the associated step is expected to advance the goal under the stated constraints. We then design a reward system that evaluates the quality of both the cognitive traces and the generated steps, while additionally rewarding higher-quality cognitive traces that improve downstream step quality. This design encourages the model to generate cognitive traces that are causally useful for downstream methodological step generation, rather than optimizing traces only as standalone explanations. During training, an external LLM-based reward model assigns rewards to both the cognitive traces and the method steps. Policy updates are performed using GRPO over these trajectory-level rewards, enabling the model to iteratively refine both its cognitive trace and method generation.

Across two models, Llama-3.2-3B-Instruct and Qwen-2.5-3B-Instruct, and two scientific domains, Artificial Intelligence and Materials Science, \textsc{CogTRL} consistently outperforms zero-shot COT prompting, supervised fine-tuning (with and without cognitive traces), and vanilla GRPO (without cognitive traces) by an average of 7.85 points.

We find that the 3B models trained using \textsc{CogTRL} match the performance of much larger (70B+ parameter) models. Moreover, in our blind human evaluation, domain experts preferred methods generated by \textsc{CogTRL} 71.42\% of the time across both the AI and the Materials Science domains. Finally, we show that \textsc{CogTRL} preserves and in some cases modestly improves the model's performance on out-of-domain reasoning tasks.

\section{Related Work}
\subsection{LLM Agents for Scientific Discovery}
LLM-based agents have accelerated scientific discovery through retrieval-augmented, literature-centric, and multi-agent workflows that leverage citation structures, explicit constraints, graph reasoning, grounded exploration, experimental planning, and autonomous execution for scientific question answering, literature synthesis, and hypothesis generation \citep{lala2023paperqa, agarwal2024litllm, wang2024scimon, baek2025researchagent, yang2024moose, kumbhar2025hypothesis, ghafarollahi2024sciagents, sprueill2024chemreasoner, bran2023chemcrow, jia2024llmatdesign, ghafarollahi2024protagents, gottweis2025towards, liu2024drugagent, yamada2025ai, ai4science2023impact}. These approaches typically rely on closed-source frontier LLMs with access to multiple external tools or simulators, limiting accessibility and cross-domain generalization. In contrast, our work focuses on improving the intrinsic scientific reasoning capabilities of open-source LLMs through training, internalizing constraint-aware and structured cognitive reasoning for open-ended scientific method generation without external tool dependence.

\subsection{Training LLMs for Scientific Discovery and Reasoning}

Prior work has explored training-based approaches for scientific discovery through domain-adaptive pretraining, Supervised Fine-Tuning (SFT), Reinforcement Learning (RL), rationale distillation, and process supervision. Pretraining and SFT improve domain-specific reasoning and generation for tasks such as scientific question answering, information extraction, and material generation using large scientific corpora and instruction datasets \citep{luo2022biogpt, beltagy2019scibert, ding2024matexpert}. RL has further been applied to optimization-driven scientific settings such as molecular design, physical system modeling, and evolutionary algorithm discovery using outcome-based rewards \citep{peng2023study, cavanagh2024smileyllama, hu2023novo, surina2025algorithm}. 

Concurrently, reasoning-focused works train LLMs using explicit or implicit reasoning traces through rationale bootstrapping, process supervision, RL-based reasoning, and latent reasoning \citep{zelikman2022star, zelikman2024quiet, hsieh2023distilling, lightman2024let, shen2025satori, wu2024thinking, hao2024training}. However, these approaches primarily target math, QA, or programmatically verifiable tasks. In contrast, \textsc{CogTRL} integrates cognitively grounded intermediate traces with RL for open-ended, constraint-aware scientific method generation, jointly rewarding both cognitive traces and downstream step quality.

\begin{figure*}[t]  
  \centering
  \includegraphics[width=\textwidth]{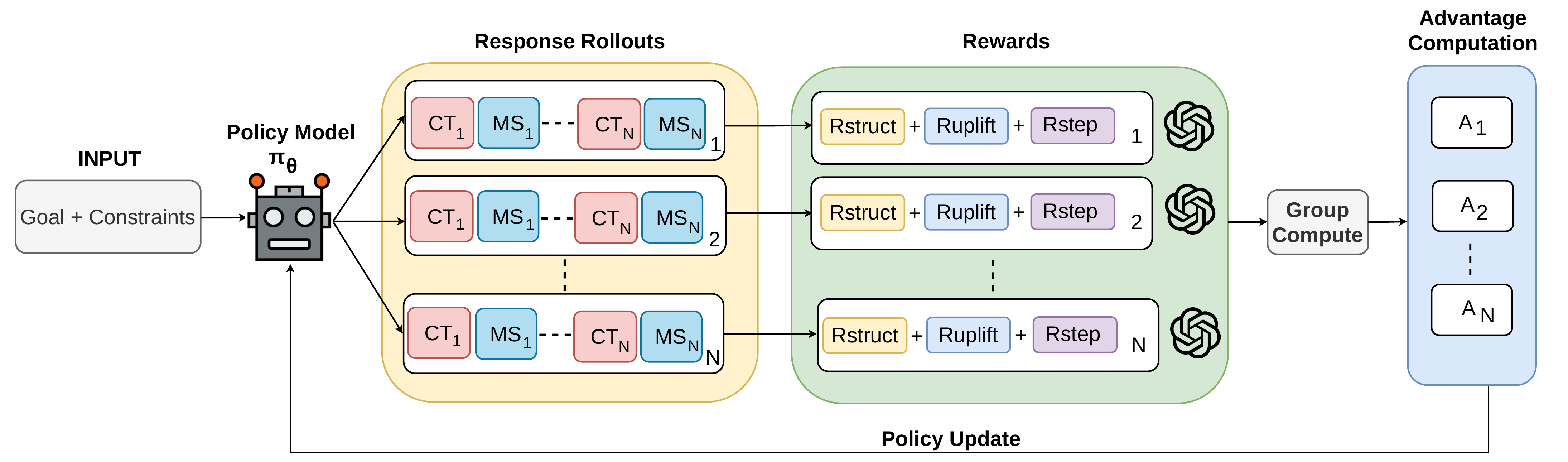}
\caption{
\textbf{Overview of \textsc{CogTRL}}: Given a goal and constraints, the policy generates rollouts containing interleaved cognitive traces $(CT_1,\dots,CT_N)$ and method steps $(MS_1,\dots,MS_N)$. The reward model computes $R_{\text{trace}}$, and $R_{\text{step}}$, which are aggregated in $R_{\text{total}}$ to estimate group-wise advantages for GRPO-style policy updates.
}
  \label{fig:main-diagram}
\end{figure*}

\section{Method}

To model goal-conditioned scientific method generation, we introduce cognitive trace and method trajectories as a formal abstraction. Given an input \( x = (g, C) \) consisting of a research goal \( g \) and a set of constraints \( C \), the LLM generates the trajectory \( \tau = (t_1, s_1, t_2, s_2, \ldots, t_n, s_n) \), where each \( t_i \) is a brief cognitive trace, and each \( s_i \) is a corresponding methodological step conditioned on previous cognitive traces and steps. Trajectories are generated auto-regressively as follows: \( \pi_{\theta}(\tau \mid x) = \prod_{i=1}^{n} \pi_{\theta}(t_i \mid x, \tau_{< i}) \, \pi_{\theta}(s_i \mid x, \tau_{\le i}) \).

\subsection{Reward Design}

Following prior work \citep{kumbhar2025hypothesis, yang2024moose} that uses LLMs as rubric-based reward models for supervision in open-ended scientific discovery, we design a trajectory-level reward that evaluates both cognitive traces (\( t_i \)) and methodological steps (\( s_i \)). We use o3-mini \citep{openai2025o3mini}, a Large Reasoning Model (LRM), as our reward model. We prompt this LRM to return a list of scores for each trace and step, which are then aggregated for policy optimization. The prompts for reward calculation are listed in Appendix \ref{Appendix: Prompts for Reward Calculation} and \ref{Appendix: CogThought Reward}. We additionally include a structural reward to enforce valid formatting. Refer to Algorithm \ref{alg:cogtrl_loopscoped}.

\subsubsection{Cognitive-Trace Quality Reward}
\label{sub-sec:cog-trace-reward}

We define a cognitive-trace quality reward, \( R_{\text{trace}} \), that evaluates whether the generated traces reflect the type of cognitive reasoning useful for scientific method generation. The LRM judge scores the traces on a scale of 1 to 5 along six dimensions:
\noindent
\textbf{(1) Goal and Constraint Integration:} Degree to which the reasoning links decisions to the research goal and constraints.
\textbf{(2) Scientific and Mechanistic Reasoning:} Use of appropriate scientific principles or mechanisms to justify the action.
\textbf{(3) Causal Logic and Actionability:} Presence of forward-looking cause–effect reasoning that directly supports the step.
\textbf{(4) Information Density:} Amount of meaningful scientific insight conveyed without unnecessary verbosity.
\textbf{(5) Scientific Accuracy and Consistency:} Factual correctness and logical coherence of the reasoning.
\textbf{(6) Uncertainty and Trade-offs:} Explicit consideration of alternatives, limitations, or uncertainties motivating the choice.

\subsubsection{Method-Step Quality Reward}
\label{Method-Step Quality Reward}

We define a method-step quality reward, \(R_{\text{step}}\), that evaluates just the generated methodological steps. Each step is scored from 1 to 5 along the six dimensions:
\noindent\textbf{(1) Alignment with Research Objectives and Constraints:}
Whether the steps directly address the research goal and adhere to the provided constraints.
\noindent\textbf{(2) Scientific Plausibility:}
Whether the steps are grounded in scientific principles and reflect realistic mechanisms.
\noindent\textbf{(3) Innovation and Novelty:}
Whether the methodology introduces creative, non-obvious, or insightful scientific strategies.
\noindent\textbf{(4) Testability:}
Whether the steps can be empirically validated or experimentally measured.
\noindent\textbf{(5) Feasibility and Scalability:}
Whether the methodology is practical to implement and can be extended or scaled.
\noindent\textbf{(6) Impact Potential:}
Whether the proposed steps meaningfully advance scientific understanding or the research objective.

\subsubsection{Uplift Reward: Encouraging Step Quality via Cognitive Traces}

Since the generated method is conditioned on the cognitive trace, we introduce a multiplicative uplift reward, $R_{\text{uplift}}$, that is responsible for optimizing the traces that are useful for generating the steps. It is defined as \( R_{\text{uplift}} = R_{\text{step}} \cdot \sigma\!\left(R_{\text{trace}} - \alpha\right) \), where rewards are normalized to \([0,1]\), \(\sigma(\cdot)\) is the sigmoid function, and \(\alpha\) is a trace-quality threshold. This formulation reinforces cognitive traces to improve the quality of the resulting step. We set \(\alpha=0.6\) for our experiments. More details in Appendix \ref{Appendix:Reward Formulation}.

\subsubsection{Structural Rewards}
\label{Structural Rewards}

We apply a structural reward, $R_{\text{struct}}$, that enforces the required trajectory format of \texttt{<Trace\_i>} \dots \texttt{</Trace\_i>} \texttt{<Step\_i>} \dots \texttt{</Step\_i>}, with correct alternation, matched tags, and consecutive indices.

\subsection{Cognitive Traces Reinforcement Learning}

We formalize \textsc{CogTRL} (Figure~ \ref{fig:main-diagram}) as optimizing a policy $\pi_\theta$ to maximize the expected total reward $R_{\text{total}}$ using Group Relative Policy Optimization (GRPO). Unlike vanilla GRPO, which optimizes over method steps alone, \textsc{CogTRL} optimizes a joint \emph{trace–step} policy, explicitly optimizing cognitive traces alongside methodological steps in an interleaved manner. At each iteration, we sample inputs, $x \sim \mathcal{D}$, and generate $G$ trace–step trajectories $\{\tau_1,\dots,\tau_G\}$ from the current policy $\pi_{\theta_{\text{old}}}$. For each trajectory, we compute $r_i = R_{\text{total}}(\tau_i)$ and define the group-normalized advantage
\[
A_i =
\frac{r_i - \mathrm{mean}(\{r_1,\dots,r_G\})}
{\mathrm{std}(\{r_1,\dots,r_G\}) + \epsilon},
\]
which reduces variance via within-group comparison. 
We update the policy using the standard clipped GRPO objective,
$\mathbb{E}[\mathcal{L}_{\mathrm{GRPO}}(\theta)]$, as:
\begin{equation}
\small
\begin{aligned}
\mathbb{E}\!\Bigg[
&\frac{1}{G}\sum_{i=1}^{G}
\min\!\Big(
\rho_i A_i,\,
\mathrm{clip}(\rho_i,1-\epsilon_c,1+\epsilon_c)A_i
\Big)
\\[-2pt]
&\quad - \beta D_{\mathrm{KL}}(\pi_\theta \| \pi_{\mathrm{ref}})
\Bigg],
\end{aligned}
\end{equation}
where $\rho_i(\theta)=\pi_\theta(\tau_i|x)/\pi_{\theta_{\text{old}}}(\tau_i|x)$ is the policy ratio. The KL term constrains deviation from the SFT reference policy $\pi_{\mathrm{ref}}$, promoting training stability. The total reward that we use for training is: \( R_{\text{total}} = R_{\text{step}} + \gamma R_{\text{uplift}} + \lambda R_{\text{struct}}. \)

\section{Experimental Setup}
We perform training experiments in two settings: (1) \emph{interleaved cognitive trace-step generation}, where reasoning traces and method steps are generated alternately, and (2) \emph{think-first generation}, similar to the current thinking LLMs, where the model first generates cognitive traces followed by the complete methodology steps. We specifically select open-source LLMs with a knowledge cut-off of 2023, as our test set consists of papers from 2024 to avoid data leakage. Refer to Appendices \ref{Appendix: Traininng Details}, \ref{Appendix: Cost} and \ref{Appendix: Prompts for ZS, SFT, RL} for more training details.

\paragraph{Baselines \& Models}
We compare \textsc{CogTRL} against the following baselines: 1) Zero-Shot Chain-of-Thought (CoT), 2) SFT without cognitive traces, 3) SFT with cognitive traces, and 4) Vanilla GRPO (no traces just steps) trained only with step-quality rewards. Zero-shot CoT is evaluated across models spanning small to frontier scale, including Llama-3.2-3B-Instruct, Qwen-2.5-3B-Instruct, Llama-3.3-70B-Instruct, Qwen-2.5-72B-Instruct, Gemini-2.5-Pro, GPT-5.4, and Claude Opus 4.6, while all experiments involving training (SFT and RL) are conducted on Llama-3.2-3B-Instruct and Qwen-2.5-3B-Instruct. We abbreviate instruction-tuned models using “It” for “Instruct” (e.g., LLaMA-3.2-3B-It and Qwen-2.5-3B-It).

\begin{table}[t]
\centering
\footnotesize
\resizebox{\columnwidth}{!}{%
\begin{tabular}{lcc}
\toprule
\textbf{Domain} & \textbf{Samples} & \textbf{Percentage (\%)} \\
\midrule
Physics (Phy)              & 1{,}311 & 26.3 \\
Computer Science (CS)      & 1{,}225 & 24.5 \\
Mathematics (Math)         & 1{,}116 & 22.4 \\
Artificial Intelligence (AI) & 583   & 11.7 \\
Electrical Engineering (EESS) & 570  & 11.4 \\
Biology (Bio)              & 185     & 3.7 \\
\midrule
\textbf{Total}             & \textbf{4{,}990} & \textbf{100.0} \\
\bottomrule
\end{tabular}%
}
\caption{Distribution of SFT training data across various domains on arxiv.}
\label{tab:sft_statistics}
\end{table}

\paragraph{Training Data}

To the best of our knowledge, a dataset containing cognitive traces for scientific discovery does not exist. We therefore construct a training dataset from arXiv papers spanning six domains: Physics, Computer Science, Mathematics, AI, Electrical Engineering, and Biology. For each paper, we use an LLM-based modular extraction pipeline consisting of Goal, Constraint, and Method Extraction agents to obtain the research goal, constraints, and step-by-step methodology.

To validate the extracted data, we randomly sample 10 examples per domain and ask four annotators to evaluate the correctness of the extracted goals, constraints, and methods against the original papers using a 3-point scale (1: low, 2: medium, 3: high). The dataset achieves an average correctness score of $2.58/3$ with Krippendorff’s alpha of $0.71$, indicating substantial agreement. Disagreements are resolved using annotator confidence labels. Since ground-truth cognitive traces are unavailable, and constructing human-written traces at scale is labour-intensive, we generate them using GPT-4o conditioned on the extracted goals, constraints, and methodologies, producing three candidates per sample, following prior work that uses LLM-generated rationales as auxiliary supervision for training smaller models \citep{zelikman2022star, hsieh2023distilling}. Final traces are selected using the highest average score assigned by GPT-4o, Gemini-2.5-Pro, and OpenAI-o1 under the taxonomy defined in Section~\ref{sub-sec:cog-trace-reward}. These teacher-generated traces are used only for SFT initialization, while RL training later explores traces and steps through policy-generated rollouts. Additional dataset construction details, prompts, and validation procedures are provided in Appendix~\ref{Appendix: SFT Data Construction}, and \ref{Prompts for SFT Data Cons}. The final dataset statistics are shown in Table~\ref{tab:sft_statistics}.

\paragraph{Evaluation Benchmark}
Our evaluation consists of two different subsets. First is a held-out test set spanning two domains: AI (in-domain) and Materials Science (out-of-domain). The AI test set includes 50 NeurIPS 2024 papers sampled across multiple subareas (e.g., LLMs, optimization, reinforcement learning), with goals, constraints, and step-by-step methodologies manually extracted by two independent domain experts. The Materials Science test set comprises 50 papers from the MatDesign benchmark \citep{kumbhar2025hypothesis}, which provides goals and constraints, while two domain experts extract methodological steps. Second, are several out-of-domain existing reasoning benchmarks (e.g., AMC23, AIME, MMLU-Pro \citep{wang2024mmlu}, GPQA-Diamond \citep{rein2024gpqa}, HumanEval \citep{chen2021evaluating}, OlympiadBench \citep{he2024olympiadbench}, etc. to measure how well the base model capabilities are preserved while undergoing training for scientific discovery assistance. To assess test-set reliability, we conduct a reference-guided expert validation study in which two domain experts per domain independently rate each extracted sample against the original paper on a 3-point ordinal scale: 1 = Low, 2 = Medium, and 3 = Highly Accurate. We report quadratic weighted Cohen's $\kappa$ to account for ordinal ratings, obtaining $\kappa=0.67$, indicating substantial inter-annotator agreement. Details are provided in Appendix~\ref{Appendix: Test Set Construction}.

\paragraph{Evaluation Metrics}
Evaluating open-ended scientific methodologies is challenging due to their multi-dimensional and domain-specific nature, while reliable verification often requires physical experiments, simulation pipelines, or domain-specific infrastructure. Prior work \citep{d2025yescieval, zhang2025sentient, xiong2025llava, bavaresco2025llms, handa2025optagent, badshah2025reference, tong2024codejudge, lan2024criticeval, tang2024llms, krolik2024towards} shows that LLMs can serve as reliable evaluators across language, code, and scientific reasoning tasks.

Following \citet{kumbhar2025hypothesis}, we evaluate only generated methodological steps, not traces, using two independent agentic evaluation frameworks: Codex \citep{openai2025codex} and Claude Code \citep{anthropic2025claudecode}, supported by GPT-5.4 and Claude Sonnet 4.6, respectively, both with high reasoning settings, internet search, and tool-use capabilities enabled in identical settings. These evaluators are distinct from the OpenAI o3-mini reward model used during RL training, reducing reward-model overfitting. During training and evaluation, we explicitly instruct the reward models not to reward stylistic verbosity, unsupported claims, hallucinated scientific content, or uncertain reasoning. Evaluation is rubric-based rather than reference-matching: judges score six methodological quality dimensions from Section~\ref{Method-Step Quality Reward}. During evaluation, retrieval is unrestricted, allowing judges to use external documents, research papers, and databases. Judges additionally provide brief score justifications. Scores are assigned independently on a 1--5 scale and aggregated across judges. Across both domains, aggregated LLM-judge evaluations align with human expert preferences in $77.14\%$ of cases. To reduce position bias and long-context effects, methodologies are evaluated independently rather than through pairwise comparisons \citep{shi2024judging,hosseini2025efficient}. All results are averaged across three independent runs with standard deviations reported for Tables~\ref{tab:combined_quality_results} and \ref{tab:trace_inference}. Additional evaluation details are provided in Appendices~\ref{Appedix: Score Scaling}, \ref{Appendix: Prompts for Reward Calculation}, \ref{Appendix: CogThought Reward}, and \ref{Appendix: Eval Judge Config}.

As an additional safeguard against reward hacking, we evaluate LLaMA-3.2-3B-It and Qwen-2.5-3B-It under zero-shot, GRPO, and \textsc{CogTRL} settings using an independent evaluation rubric \citep{chen2026mlr}. \textsc{CogTRL} consistently outperforms the baselines under this alternative rubric as shown in Appendix \ref{Appendix:Independent Rubric Evaluation}.

\begin{table}[t]
\centering
\resizebox{\columnwidth}{!}{%
\Large
\begin{tabular}{@{} l l c c @{}}
\toprule
\textbf{Model} & \textbf{Technique} &
\makecell{\textbf{Quality}\\\textbf{AI}} &
\makecell{\textbf{Quality}\\\textbf{MatSci}} \\
\midrule

Llama-3.2-3B-It 
& Zero-shot-CoT (Think First)         
& 37.30 $\pm$ 0.92
& 41.98 $\pm$ 0.88 \\

& SFT with traces (Think First) 
& 44.27 $\pm$ 0.79
& 42.33 $\pm$ 0.83 \\

& \textsc{CogTRL} (Think First)            
& 48.20 $\pm$ 0.68
& 49.31 $\pm$ 0.64 \\

& Zero-shot-CoT (Int. Think)         
& 38.00 $\pm$ 0.90
& 42.46 $\pm$ 0.86 \\

& SFT w/o traces  
& 42.44 $\pm$ 0.81
& 45.77 $\pm$ 0.73 \\

& SFT with traces (Int. Think) 
& 45.84 $\pm$ 0.71
& 47.40 $\pm$ 0.69 \\

& GRPO (Int. Think)              
& 48.47 $\pm$ 0.66
& 50.37 $\pm$ 0.61 \\

& \textbf{\textbf{\textsc{CogTRL}} (Int. Think)}            
& \textbf{52.63 $\pm$ 0.58}
& \textbf{55.64 $\pm$ 0.51} \\
\midrule

Qwen-2.5-3B-It  
& Zero-shot-CoT (Think First)         
& 39.61 $\pm$ 0.87
& 42.62 $\pm$ 0.85 \\

& SFT with traces (Think First) 
& 42.84 $\pm$ 0.80
& 43.36 $\pm$ 0.82 \\

& \textsc{CogTRL} (Think First)            
& 46.97 $\pm$ 0.70
& 49.51 $\pm$ 0.66 \\

& Zero-shot-CoT (Int. Think)         
& 40.50 $\pm$ 0.84
& 44.03 $\pm$ 0.80 \\

& SFT w/o traces   
& 43.13 $\pm$ 0.76
& 46.20 $\pm$ 0.71 \\

& SFT with traces (Int. Think) 
& 45.10 $\pm$ 0.72
& 47.70 $\pm$ 0.67 \\

& GRPO (Int. Think)              
& 47.24 $\pm$ 0.69
& 50.03 $\pm$ 0.63 \\

& \textbf{CogTRL (Int. Think)}            
& \textbf{51.13 $\pm$ 0.61}
& \textbf{53.17 $\pm$ 0.56} \\
\midrule

Llama-3.3-70B-It 
& Zero-shot-CoT (Int. Think)       
& 52.27 $\pm$ 0.57
& 54.00 $\pm$ 0.53 \\

Qwen-2.5-72B-It 
& Zero-shot-CoT (Int. Think)       
& 50.47 $\pm$ 0.62
& 54.57 $\pm$ 0.49 \\

GPT-5.4         
& Zero-shot-CoT (Int. Think)    
& 60.27 $\pm$ 0.42   
& \textbf{64.60 $\pm$ 0.37} \\

Gemini-2.5-Pro         
& Zero-shot-CoT (Int. Think)   
& 58.64 $\pm$ 0.47   
& 59.10 $\pm$ 0.44 \\

Claude Opus 4.6        
& Zero-shot-CoT (Int. Think)    
& \textbf{63.87 $\pm$ 0.35}
& 62.53 $\pm$ 0.41 \\

\bottomrule
\end{tabular}%
}
\caption{
Quality results for \textsc{CogTRL} and baselines across two generation formats: \emph{Think First}, where cognitive traces are generated before the methodology, and \emph{Int. Think}, where traces are interleaved with method.
}
\label{tab:combined_quality_results}

\end{table}

\section{Results and Analysis}
\paragraph{\textsc{CogTRL} with Interleaved Trace-Step generation improves Quality of Scientific Methods}

Table~\ref{tab:combined_quality_results} shows that \textsc{CogTRL} outperforms baselines across both domains and generation settings. In the interleaved setting, \textsc{CogTRL} improves over GRPO by an average of $4.12$ points and over all non-CogTRL baselines using the same 3B models by $7.85$ points. The 3B \textsc{CogTRL} models become competitive with open source 70B models, outperforming Qwen-2.5-72B-Instruct on AI and matching or exceeding Llama-3.3-70B-Instruct.

Interleaved reasoning consistently outperforms think-first generation. Averaged across both models, interleaved \textsc{CogTRL} improves over think-first \textsc{CogTRL} by $4.30$ points on AI and $4.99$ on Material Science. Similar trends for SFT with traces suggest that conditioning each step on preceding traces and steps yields stronger alignment than generating all reasoning upfront.

We further study the contribution of the uplift reward $R_{\text{uplift}}$. Removing it while retaining $R_{\text{step}}$, $R_{\text{trace}}$, and $R_{\text{struct}}$ causes an average degradation of $5.38\%$ across both models (Table~\ref{tab:Reward upliftment ablation}), suggesting that cognitive traces are most beneficial when they improve downstream method quality and not just superficial explanations. Overall, combining RL with cognitive-trace and multidimensional quality rewards enables high-quality method generation, with similar gains observed using Dr. GRPO \citep{liu2025understanding} (Appendix \ref{Appendix: Dr.GRPO} Table~\ref{tab:CoGTRL with DrGRPO}), suggesting that \textsc{CogTRL} is not tied to a specific RL algorithm.

\paragraph{Uplift Evidence of Cognitive Traces}
To evaluate whether cognitive traces improve scientific step generation quality, we compute teacher-forced per-token probabilities for fixed \textsc{CogTRL}-generated methodological steps under three settings: aligned traces (w/A), no traces (w/O), and mismatched traces from another example but same domain (w/M). “High-quality steps” refer to \textsc{CogTRL} generations evaluated using the aggregate LLM-judge quality scores; no additional filtering or reranking is performed for this analysis. We score only the \texttt{Step\_i} tokens and report $\exp(-\mathrm{NLL})$, where NLL (Negative Log Likelihood) is averaged over the same target step tokens across all conditions, preventing longer trace inputs from artificially improving scores. As shown in Table~\ref{tab:trace_logprob}, aligned traces consistently yield higher probabilities than both the no-trace and mismatched-trace settings across models and domains. While the w/O condition yields lower probabilities for high-quality steps, the w/M condition demonstrates that the gains arise from trace--step alignment rather than additional context length or formatting.
Further, Table~\ref{tab:trace_inference} shows that \textsc{CogTRL} models consistently generate higher-quality methodologies when prompted to generate cognitive traces during inference, supporting that traces improve downstream method quality.

\begin{table}[t]
\centering
\resizebox{\columnwidth}{!}{%
\Large
\begin{tabular}{@{} l c c @{}}
\toprule
\textbf{Model} &
\makecell{\textbf{Quality}\\\textbf{AI}} &
\makecell{\textbf{Quality}\\\textbf{MatSci}} \\
\midrule

\textsc{CogTRL} Llama-3.2-3B-It w/o uplift & 50.65 & 53.20 \\
\textsc{CogTRL} Qwen-2.5-3B-It w/o uplift  & 48.35 & 48.96 \\

\bottomrule
\end{tabular}%
}
\caption{\textbf{Ablation without Reward Upliftment}: Removing $R_{\text{uplift}}$ from the cumulative reward degrades performance across both models.}
\label{tab:Reward upliftment ablation}
\end{table}

\begin{figure*}[t]
  \centering
  \includegraphics[width=\textwidth]{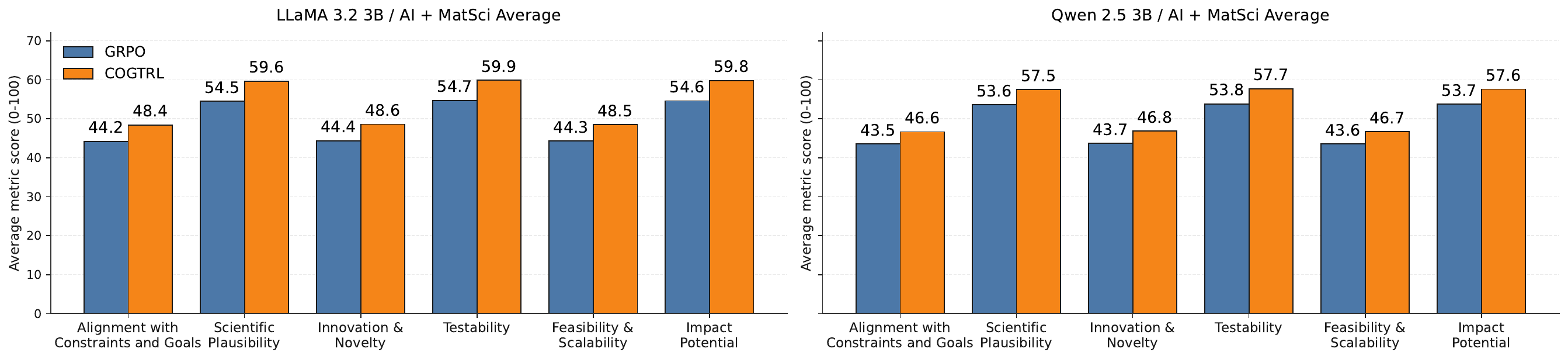}
  \phantomsection
  \caption{The above plots show the scores for every dimension of the Step Quality metric achieved by models trained with GRPO and \textsc{CogTRL}. Both models, Llama-3.2-3B-Instruct (Left) and Qwen-2.5-3B-Instruct (Right), trained with \textsc{CogTRL} perform better on all dimensions compared to GRPO.}
  \label{fig:micro-quality-improvements}
\end{figure*}

\paragraph{Improvements Across Sub-Metrics}
To identify where \textsc{CogTRL} provides the greatest benefits, we compute average per-dimension improvements across Llama-3.2-3B-Instruct and Qwen2.5-3B-Instruct. Averaged across models and domains, \textsc{CogTRL} yields the largest gains in \textit{Testability} (+4.55) and \textit{Impact Potential} (+4.55), followed by \textit{Scientific Plausibility} (+4.50). More moderate improvements are observed for \textit{Alignment with Constraints and Goals} (+3.65), \textit{Innovation and Novelty} (+3.65), and \textit{Feasibility and Scalability} (+3.65). Figure~\ref{fig:micro-quality-improvements} reports the per-dimension breakdown relative to the vanilla GRPO baseline. These results indicate that \textsc{CogTRL} delivers balanced improvements across quality dimensions.




\begin{table}[t]
  \centering
  \small
  \setlength{\tabcolsep}{0pt}
  \renewcommand{\arraystretch}{1.08}

  \begin{tabular*}{\columnwidth}{@{\extracolsep{\fill}}lcccc@{}}
  \toprule
  \textbf{Model} & \textbf{Dom.} & \cellcolor{blue!12}\textbf{w/A} & \textbf{w/O} & \textbf{w/M} \\
  \midrule
  Llama 3.2 3B-It & AI     & \cellcolor{blue!12}0.6324 & 0.5274 & 0.4732 \\
  Llama 3.2 3B-It & MatSci & \cellcolor{blue!12}0.6108 & 0.5145 & 0.4598 \\
  Qwen 2.5 3B-It  & AI     & \cellcolor{blue!12}0.5028 & 0.4127 & 0.3631 \\
  Qwen 2.5 3B-It  & MatSci & \cellcolor{blue!12}0.4859 & 0.3991 & 0.3473 \\
  \bottomrule
  \end{tabular*}

  \caption{Per-token probabilities of fixed high-quality methodological steps under aligned traces (w/A), no traces (w/O), and mismatched traces (w/M).}
  \label{tab:trace_logprob}
\end{table}

\begin{table}[t]
\centering
\small
\setlength{\tabcolsep}{2.3pt}
\renewcommand{\arraystretch}{0.95}

\begin{tabularx}{\columnwidth}{@{}Xlccc@{}}
\toprule
\textbf{Model} & \textbf{Method} & \textbf{Traces} & \textbf{AI} & \textbf{MatSci} \\
\midrule
\multirow{3}{=}{Llama 3.2 3B-It}
& GRPO & --  & $48.47{\pm}1.12$ & $50.37{\pm}1.08$ \\
& \textsc{CogTRL} & w/o & $49.50{\pm}1.05$ & $51.12{\pm}1.14$ \\
& \textbf{\textsc{CogTRL}} & \textbf{w/} & $\mathbf{52.63{\pm}1.01}$ & $\mathbf{55.64{\pm}1.09}$ \\
\midrule
\multirow{3}{=}{Qwen 2.5 3B-It}
& GRPO & --  & $47.24{\pm}1.18$ & $50.03{\pm}1.10$ \\
& \textsc{CogTRL} & w/o & $47.65{\pm}1.11$ & $52.05{\pm}1.07$ \\
& \textbf{\textsc{CogTRL}} & \textbf{w/} & $\mathbf{51.13{\pm}1.04}$ & $\mathbf{53.17{\pm}1.12}$ \\
\bottomrule
\end{tabularx}
\caption{Impact of cognitive traces during inference. Including traces during inference consistently improves methodology quality across models and domains.
\label{tab:trace_inference}
}
\end{table}

\paragraph{Impact of Supervised Fine-Tuning}
Table~\ref{tab:combined_quality_results} shows that SFT with cognitive traces consistently improves scientific methodology quality over zero-shot prompting and SFT without traces across AI and Materials Science. Since SFT without traces does not generate cognitive reasoning, no think-first or interleaved setting applies to this baseline. In contrast, interleaved SFT with traces consistently outperforms think-first SFT with traces across both 3B models and domains, suggesting that conditioning each step on preceding traces improves methodological alignment. These results suggest that cognitive traces provide useful intermediate reasoning signals, while SFT-with-traces initialization further improves \textsc{CogTRL} performance.

\begin{table*}[t]
\centering
\small
\resizebox{\textwidth}{!}{
\begin{tabular}{l l c c c c c c c}
\toprule
\textbf{Model} & \textbf{Method} & \textbf{AIME} & \textbf{AMC23} & \textbf{GPQA} & \textbf{HumanEval} & \textbf{MMLU-Pro} & \textbf{Olympiad} & \textbf{Avg.} \\
\midrule
\multirow{5}{*}{Llama-3.2-3B-It} 
& Zero Shot          & 6.46 & 15.83 & 24.75 & \textbf{40.24} & 31.00 & 8.17 & 21.07 \\
& SFT w/o traces   & 4.97 & 8.33 & 17.68 & 26.63 & 16.43 & 4.29 & 13.05 \\
& SFT with traces  & 7.50 & 14.17 & 17.68 & 6.30 & 11.12 & 5.56 & 10.38 \\
& GRPO               & \textbf{8.22} & 11.67 & \textbf{25.08} & 39.84 & \textbf{31.61} & 7.98 & 20.73 \\
& CogTRL             & 8.15 & \textbf{17.50} & 24.24 & 38.41 & 31.51 & \textbf{8.46} & \textbf{21.37} \\
\midrule
\multirow{5}{*}{Qwen-2.5-3B It} 
& Zero Shot          & 12.00 & 34.17 & 29.46 & 50.81 & 41.92 & \textbf{23.72} & 32.01 \\
& SFT w/o traces   & 3.72 & 19.17 & 27.44 & 21.75 & 34.91 & 12.90 & 19.98 \\
& SFT with traces  & 5.21 & 23.33 & 26.43 & 31.91 & 33.57 & 14.44 & 22.48 \\
& GRPO               & \textbf{13.11} & 39.17 & 27.44 & 50.00 & 41.93 & 22.35 & 32.33 \\
& CogTRL             & 11.47 & \textbf{40.00} & \textbf{29.80} & \textbf{54.27} & \textbf{42.57} & 21.76 & \textbf{33.31} \\
\bottomrule
\end{tabular}
}
\caption{Performance on general-purpose reasoning benchmarks under interleaved thinking averaged across three runs. Best scores per model and benchmark are shown in bold. \textsc{CogTRL} preserves overall reasoning capability.}
\label{tab:general_benchmarks}
\end{table*}

\paragraph{Out-of-Domain Performance of \textsc{CogTRL}}
Table~\ref{tab:general_benchmarks} shows that \textsc{CogTRL} preserves the general reasoning capabilities of the base instruction-tuned models across math (AIME, AMC23), scientific (GPQA-Diamond, OlympiadBench), code generation (HumanEval), and language understanding (MMLU-Pro) tasks. Unlike SFT, which causes degradation across several benchmarks, \textsc{CogTRL} avoids this collapse while improving scientific method generation. Relative to vanilla GRPO, \textsc{CogTRL} achieves modest gains on benchmarks such as AMC23 and GPQA-Diamond, with only minor regressions on others. Importantly, average benchmark performance remains stable across both model families, indicating that gains in scientific discovery assistance do not come at the expense of broad reasoning competence.

\subsection{Human Evaluation}
We conduct a human expert preference evaluation with four annotators, two Chemistry PhD students, one CS PhD student, and one CS Master’s student, on AI and Materials Science methodologies. Across all 50 samples per domain and three techniques (vanilla Instruct, GRPO, and \textsc{CogTRL}), resulting in 300 total evaluations, randomized methodology outputs are independently ranked by two domain experts with confidence scores; disagreements are resolved using the higher-confidence labels. The inter-annotator agreement measured using Kendall’s W is $0.70$, indicating substantial agreement. Across all pairwise comparisons, \textsc{CogTRL} is preferred in $71.42\%$ of cases, with annotators consistently highlighting superior completeness, organization, and causal structure.

In Materials Science, experts attribute these gains to greater specificity and actionability, including clearer synthesis procedures, characterization workflows, and appropriate use of tools such as molecular dynamics and density functional theory, along with stronger integration of constraints and deployment context (e.g., environmental conditions, biofouling, safety, and deployment scenarios). They also note more systematic workflows from material selection to validation. In AI, experts report that \textsc{CogTRL} better grounds methodologies in conceptual reasoning before translating them into concrete training and evaluation procedures, while more explicitly addressing constraints such as domain adaptation and verification, resulting in a systematic and comprehensive methodology.

\paragraph{Case Study: Constraint-Driven Design for Photothermal Self-Healing Elastomers.}

We compare three model-generated methodologies for designing a sustainable photothermal-responsive self-healing elastomer under three coupled constraints: (i) high mechanical strength with repeatable healing, (ii) balanced photothermal efficiency and dispersion, and (iii) fast localized NIR-triggered repair. The zero-shot model proposes a generic workflow but fails to explicitly connect network architecture, photothermal response, and healing behavior to these constraints. The GRPO model introduces sophisticated components, yet lacks a causal structure linking material choices to downstream trade-offs.

In contrast, the \textsc{CogTRL}-trained model decomposes the task into constraint-driven design stages. It jointly designs photothermal-responsive monomers and incorporation strategies for dispersion-controlled NIR absorption, explicitly analyzing morphology-dependent heating uniformity. Mechanical strength and healing are co-designed through hierarchical crosslinked networks validated under cyclic loading, while localized repair is addressed by optimizing photothermal agent geometry and thermal diffusion. Crucially, synthesis, characterization, and modeling steps are causally linked to specific constraints. Full details, along with an AI case study and actual method comparisons, are provided in Appendix~\ref{Appendix: Case Study} and \ref{Appedix:Actual Example}.

\section{Conclusion}
We presented \textsc{CogTRL}, a trajectory-level reinforcement learning framework for training LLMs to jointly generate cognitive traces and scientific methodologies for open-ended scientific discovery assistance. By optimizing cognitively grounded trace--step trajectories with multidimensional rewards, \textsc{CogTRL} improves scientific method generation across AI and Materials Science while preserving general reasoning capabilities. Experiments across multiple baselines, ablations, and human evaluations show that cognitive traces are most beneficial when they causally improve downstream methodological quality rather than serving as standalone explanations. Moreover, 3B-parameter models trained with \textsc{CogTRL} achieve competitive performance relative to the larger 70B+ models.

\section*{Limitations}
Our current study relies on state-of-the-art closed-source LRMs such as OpenAI o3-mini for reward calculation. While these models demonstrate strong scientific reasoning capabilities, they introduce additional API cost. The framework currently depends on rubric-based LLM rewards rather than physical-world, tool-based, or simulation-grounded verification, since open-ended scientific method generation lacks scalable infrastructure for explicitly validating and rewarding every generated trajectory. Although we mitigate reward hacking by separating the training reward model from the evaluation frameworks, using independent agentic evaluators with retrieval and tool-use capabilities, and explicitly discouraging hallucinated or stylistic reasoning during reward computation, the framework may still inherit biases associated with LLM-as-a-judge evaluation. We also acknowledge that rubric-based LLM evaluation might not perfectly capture real-world scientific validity. Training open-source reward models and collecting expert-authored cognitive traces remain important future directions but are beyond the scope of the current study. Finally, we experiment primarily with small-parameter models ($\approx$3B) due to compute constraints and evaluate only AI and Materials Science domains; extending \textsc{CogTRL} to larger models, broader scientific domains remains future work.

\section*{Ethics Statement}
We have utilized AI assistants, specifically Grammarly and ChatGPT, to correct grammatical errors and rephrase sentences.

\section*{Acknowledgement}
We thank the anonymous reviewers for their constructive suggestions. We extend our gratitude to the Research Computing (RC), and Enterprise Technology at Arizona State University for providing computing resources, and access to the ChatGPT enterprise version for experiments.

This research was supported by the Engineering Research and Development Center - Information Technology Laboratory (ERDC-ITL) under Contract No. W912HZ24C0022. Any opinions, findings and conclusions or recommendations expressed in this work are those of the author(s) and do not necessarily reflect the views of the ERDC-ITL. Baral was also partially supported by a DARPA contract.

\bibliography{custom}

\clearpage
\appendix
\section*{Appendix}

\begin{algorithm}[t]
\caption{\textsc{CogTRL}}
\label{alg:cogtrl_loopscoped}
\begin{algorithmic}[1]
  \small

  \Require initial policy $\pi_\theta$ initialized from the corresponding SFT-with-traces checkpoint, frozen reference policy $\pi_{\mathrm{ref}}$ copied from the same checkpoint, judges $J=(J_{\text{step}},J_{\text{trace}})$, prompt set $\mathcal{D}$, rollouts per prompt $G$, batch size $B$, hyperparameters $(\epsilon,\beta,\gamma,\lambda,\alpha)$, optimizer
  \Ensure trained policy $\pi_\theta$ (trace--step generation at train time; steps-only at inference)

  \Statex
  \For{each training iteration}
    \State Sample minibatch of prompts $\{x_j\}_{j=1}^B \sim \mathcal{D}$

    \Statex
    \For{each prompt $x_j$}
      \For{$i = 1,\dots,G$}
        \State Sample trajectory $\tau_i \sim \pi_{\theta_{\mathrm{old}}}(\cdot \mid x_j)$
        \State Parse $\tau_i \mapsto \{(t_{i,m}, s_{i,m})\}_{m=1}^{n_i}$ \Comment{strict tag parsing}
        \State $S_i \leftarrow \mathrm{concat}(\{s_{i,m}\}_{m=1}^{n_i})$ \Comment{steps only}
        \State $T_i \leftarrow \mathrm{concat}(\{t_{i,m}, s_{i,m}\}_{m=1}^{n_i})$ \Comment{full trajectory}
      \EndFor
    \EndFor

    \Statex
    \Comment{Compute per-trajectory judgments and shaped rewards}
    \For{each prompt $x_j$}
      \For{$i = 1,\dots,G$}
        \State $R_{\text{step}}(\tau_i) \leftarrow J_{\text{step}}(S_i)$
        \State $R_{\text{trace}}(\tau_i) \leftarrow J_{\text{trace}}(T_i)$
        \State $R_{\text{uplift}}(\tau_i) \leftarrow R_{\text{step}}(\tau_i)\cdot \sigma\!\big(R_{\text{trace}}(\tau_i)-\alpha\big)$
        \State $R_{\text{struct}}(\tau_i) \leftarrow \mathrm{Struct}(T_i,S_i) \in [0,1]$
        \State $R_{\text{total}}(\tau_i) \leftarrow R_{\text{step}}(\tau_i) + \gamma R_{\text{uplift}}(\tau_i) + \lambda R_{\text{struct}}(\tau_i)$
      \EndFor
    \EndFor

    \Statex
    \Comment{Group-normalized advantages (per prompt)}
    \For{each prompt $x_j$}
      \State $\mu_j \leftarrow \dfrac{1}{G}\sum_{i=1}^{G} R_{\text{total}}(\tau_i)$
      \State $\sigma_j \leftarrow \mathrm{std}\!\left(\{R_{\text{total}}(\tau_1),\dots,R_{\text{total}}(\tau_G)\}\right)$
      \For{$i = 1,\dots,G$}
        \State $\hat A_i \leftarrow \dfrac{R_{\text{total}}(\tau_i)-\mu_j}{\sigma_j+\epsilon}$
      \EndFor
    \EndFor

    \Statex
    \Comment{Policy update (GRPO-style)}
    \State Update $\pi_\theta$ using the clipped GRPO objective with advantages $\hat A_i$ and KL regularization $\beta\,D_{\mathrm{KL}}(\pi_\theta \| \pi_{\mathrm{ref}})$, where
    \Statex \hspace{1.6em} $\rho_i(\theta)=\dfrac{\pi_\theta(\tau_i \mid x_j)}{\pi_{\theta_{\mathrm{old}}}(\tau_i \mid x_j)}$
    \State $\pi_{\mathrm{old}} \leftarrow \pi_\theta$
  \EndFor

  \Statex
  \State \Return $\pi_\theta$
\end{algorithmic}
\end{algorithm}

\section{Quality \textsc{CogTRL} with Dr.GRPO}
\label{Appendix: Dr.GRPO}
Table~\ref{tab:CoGTRL with DrGRPO} evaluates \textsc{CogTRL} using Dr.GRPO as the backbone RL algorithm. Across both trained models and domains, \textsc{CogTRL} combined with Dr.GRPO consistently outperforms the vanilla Dr.GRPO baseline. For example, LLaMA 3.2 3B Instruct improves from $49.60/49.55$ to $53.15/54.80$ on AI/MatSci, while Qwen 2.5 3B Instruct improves from $49.65/50.50$ to $52.85/52.90$. These results suggest that the gains from \textsc{CogTRL} are not tied to a specific RL algorithm, and that incorporating cognitive-trace optimization improves the effectiveness of vanilla RL methods for open-ended scientific method generation.


\begin{table}[t]
\centering
\resizebox{\columnwidth}{!}{%
\Large
\begin{tabular}{@{} l l c c @{}}
\toprule
\textbf{Model} & \textbf{Technique} &
\makecell{\textbf{Quality}\\\textbf{AI}} &
\makecell{\textbf{Quality}\\\textbf{MatSci}} \\
\midrule

LLaMA 3.2 3B Instruct & Dr GRPO   & 49.60 & 49.55 \\
                      & CoGTRL w/ Dr GRPO    & 53.15 & 54.80 \\
\midrule

Qwen 2.5 3B Instruct  & Dr GRPO   & 49.65 & 50.50 \\
                      & CoGTRL w/ Dr GRPO    & 52.85 & 52.90 \\
\midrule

\end{tabular}%
}
\caption{\textbf{Performance of \textsc{CogTRL} with Dr GRPO as the RL algorithm.} \textsc{CogTRL} combined with Dr GRPO consistently outperforms the vanilla variant across domains.}
\label{tab:CoGTRL with DrGRPO}
\end{table}

\section{SFT Data Construction}
We curate a dataset spanning six domains, Physics, Computer Science, Mathematics, AI, Electrical Engineering, and Biology, where each sample contains a research goal, constraints, and a step-by-step methodology. The dataset is constructed by randomly sampling 2K arXiv paper titles per domain via the official arXiv API, parsing each paper’s HTML with BeautifulSoup, extracting its section hierarchy into a structured JSON format, and applying an LLM-based agentic pipeline to extract goals, constraints, and procedural methods for training. Specifically, papers were discarded if (i) an HTML version was unavailable (i.e., only a PDF version was provided), as direct PDF extraction was unreliable and inconsistent, or (ii) the HTML structure was malformed or incompatible with automated parsing (e.g., BeautifulSoup extraction failures). These exclusions were based solely on structural format issues and not on paper quality or domain content.  Resulting in 4990 samples after filtering. Then we randomly sample 10 extracted examples from each domain and four annotators. 
\label{Appendix: SFT Data Construction}
\subsection{Goals, Constraints and Step-by-Step Method Extraction}
From the structured JSON representation of each paper, we extract research goals, constraints, and step-by-step methods using a modular LLM-based agent pipeline. A Section Identification Agent first assigns sections and subsections to semantic categories (\textit{Goals}, \textit{Constraints}, \textit{Step-by-Step Methodology}) based on it and task-specific instructions; the corresponding content is then extracted via a python script. Three specialized agents, Goal Extraction, Constraint Extraction, and Methodology Extraction, operate only on their assigned sections to produce a concise goal statement, a structured constraint list, and an ordered, actionable procedure, respectively. Each agent uses component-specific prompts and is powered by GPT-4o 
\begin{figure*}[t]  
  \centering
  \includegraphics[width=\textwidth]{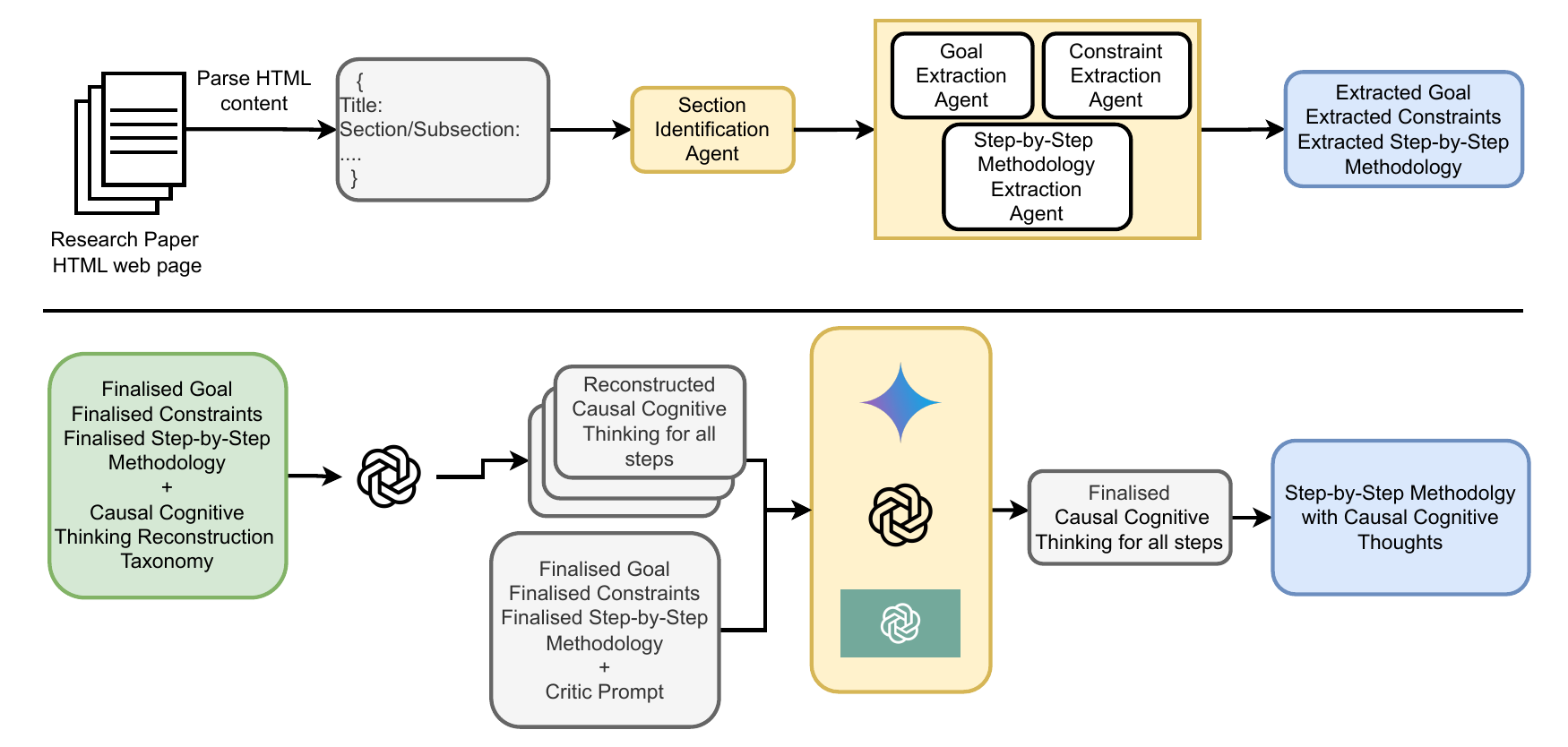}
  \caption{Overview of the four stage pipeline used in our approach.
\textbf{Stage 1}: LLM agents extract goals, constraints, and step by step methodologies from research papers and store them as structured JSON.
\textbf{Stage 2}: The extracted data is passed to GPT-4o to generate multiple candidate thoughts (T = 0.3, 0.5, 0.7). A multi-model critic system (GPT-4o, Gemini-2.5-Pro, and OpenAI-o1) selects the best thought per step.
\textbf{Stage 3}: fine-tuning is performed using only the step-by-step methodology.
\textbf{Stage 4}: fine-tuning is repeated using both the cognitive thoughts and the methodology.}
  \label{fig:wide-diagram}
\end{figure*}

\subsubsection{Cognitive Thought Generation}
We prompt GPT-4o to generate cognitive traces conditioned on extracted Goals, Constraints and Step-by-Step Methods, which gives us intermediate cognitive traces for each step, serving as annotations for SFT with thoughts experiments. To promote reasoning diversity, we sample three candidates per step with temperature values of 0.3, 0.5, and 0.7. We then implement a multi-agent critic ensemble of three independent LLMs \citep{hurst2024gpt,team2024gemini,anthropic_claude3_model_card}, where each critic scores the thought candidates on a scale of 1-5 across the six criteria mentioned in section \ref{sub-sec:cog-trace-reward}. The candidate with the maximum score is incorporated into SFT dataset.    

\section{Score Scaling and Evaluation Rubric}
\label{Appedix: Score Scaling}
Each quality dimension is scored independently on a 1--5 scale, refer Appendix \ref{Appendix: Prompts for Reward Calculation} for exact scale. The six evaluated dimensions are: Alignment with Research Objectives and Constraints, Scientific Plausibility, Innovation and Novelty, Testability, Feasibility and Scalability, and Impact Potential. For each sample, we sum the six dimension scores, divide by the maximum possible score of 30, and multiply by 100:
\[
\text{Final Score} = \frac{\sum_{i=1}^{6} s_i}{30} \times 100
\]
Judges are additionally instructed to assign the lowest plausible score when uncertain and reward only concrete, verifiable methodological detail.

\section{Test Set Construction}
\label{Appendix: Test Set Construction}
\subsection{Instructions for Human Annotators}
For each paper, annotators were asked to extract the paper's Goal, describing what the paper attempted to achieve. They were explicitly instructed not to reveal or leak information about the paper's solution to avoid introducing bias. For constraint extraction, annotators extracted all obstacles mentioned in the paper that the method was developed to address, with similar precautions to prevent solution leakage. For the step-by-step methodology, annotators extracted the detailed procedural steps described in the paper that were used to satisfy the stated goals and constraints. The annotators were told beforehand that no monetary compensation would be given, and participation is voluntary. 

\subsection{Annotator Agreements}
We ask the annotators to rate each sample on a scale of 1-3, where 1 is low, 2 is medium, and 3 is highly accurate. We calculate three scores, which are Pearson, Spearman, and Weighted Cohen Kappa with values 0.7354, 0.6724, and 0.6736, respectively. The weighted Cohen's Kappa of 0.67 indicates substantial inter-rater agreement. The Pearson correlation of 0.74 demonstrates good linear agreement between annotators, while the Spearman correlation of 0.67 shows acceptable rank-order consistency. Given the inherently subjective nature of this annotation task, which requires annotators to evaluate whether extracted goals, constraints, and step-by-step methodologies accurately capture the author's intended problem formulation and solution approach, these metrics reflect reasonable inter-rater reliability. The task involves nuanced judgments about semantic completeness, relevance, and faithful representation of research content, where some degree of interpretive variation among annotators is expected. The quadratic weighting in Cohen's Kappa appropriately accounts for the ordinal nature of the rating scale and penalizes larger disagreements more heavily, making it particularly suitable for assessing agreement on this type of quality evaluation task.

\section{Prompts for Zero Shot COT, SFT, and RL training}
\label{Appendix: Prompts for ZS, SFT, RL}
\subsection{Zero Shot COT Prompt}
\label{Appendix: Zero-Shot-CoT Prompts}
\begin{tcolorbox}[colback=black!5, colframe=black, title=System Prompt]
You are an expert research methodology consultant specializing in developing comprehensive, scientifically rigorous research approaches. Your primary function is to analyze research problems and constraints, then provide detailed, step-by-step methodological frameworks that address the specific requirements and limitations of each study. \newline

Core Responsibilities: \newline
- Understand the research problem, objectives, and constraints thoroughly \newline
- Provide actionable, detailed steps that researchers can follow \newline
- Ensure methodological rigor and feasibility within given constraints \newline
- Address potential challenges and provide solutions \newline
\end{tcolorbox}

\begin{tcolorbox}
    Response Structure Requirement: \newline
Generate the methodology as detailed sequential steps with all intermediate actions included. Each step must follow the format: \newline

<Step\_1>...</Step\_1> \newline

<Step\_2>...</Step\_2> \newline

... \newline

<Step\_n>...</Step\_n> \newline
\end{tcolorbox}

\begin{tcolorbox}[colback=black!5, colframe=black, title=Instruction Prompt]
You are given a scientific research problem.\newline
Goal:
\newline
Develop a scalable extrinsic self-healing coating system for corrosion protection of metallic structures in offshore environments. 
\newline\newline
Constraints:
\newline
1) The material should incorporate a self-healing mechanism triggered by a simple environmental factor (e.g., water).
\newline
2) The self-healing material should allow multiple healing events.
\newline
3) The coating must maintain its structural integrity and protective capabilities even after mechanical damage.
\newline
4) The material should be compatible with scalable application techniques.
\newline
5) The healing mechanism should not rely on complex multi-component reactions but rather a single-component system
\end{tcolorbox}

\begin{tcolorbox}[colback=black!5, colframe=black]
Task: Propose a comprehensive Step-by-Step Methodology to achieve the research goal while addressing the listed constraints. \newline
- Organize the answer into high-level steps (Step 1, Step 2, …). \newline
- Within each step, include detailed intermediate steps (Intermediate step 1, Intermediate step 2, …) that describe what actions should be taken, how they should be performed, and with what conditions, resources, or parameters. \newline
- Where appropriate, include scientific equations, mathematical formulations, proportions, datasets, measurements, or any other technical specifications necessary to make the methodology rigorous and directly implementable. \newline
- Ensure the methodology is precise, technically detailed, and structured so that it could be directly implemented in any relevant scientific or technical setting.
[Step\_by\_Step\_Methodology]:
\end{tcolorbox}

\subsection{SFT without Cognitive Traces}
\label{Appendix: Prompts for SFT without Cognitive Traces}
For input and system prompt we use the exact prompts used in  \ref{Appendix: Zero-Shot-CoT Prompts} as system and input instructions to the models. Below shown is a sample output for SFT without Cognitive Traces

\begin{tcolorbox}[colback=black!5, colframe=black, title=An example of Output]
Step 1: Synthesis of Silyl Ester Healing Agent \newline
- Intermediate step 1: Mix oleic acid (OA), triethylamine (TEA), and extra dry toluene in a 1:0.48:5.23 (v/v/v) ratio under an inert nitrogen atmosphere using magnetic stirring. \newline
- Intermediate step 2: Immerse the resultant mixture into an ice bath. \newline
- Intermediate step 3: Prepare a solution of chloro(dimethyl)octylsilane in extra dry toluene with a 1:0.75 (v/v) ratio. \newline
....\newline
Step 2: Preparation of Core-Shell Fibres \newline
- Prepare the core polymeric solution by mechanically dispersing 15\% synthesized silyl ester powder into a mixture of PVP and extra dry DMF (1:1.5 v/v) for 1 hour at ambient conditions. \newline
- Intermediate step 1: Measure the viscosity (~700–800 cP) and electrical conductivity (~80 $\mu$S/cm) of the core solution. \newline
...\newline
Step 3: Formulation of Corrosion Protective Coating \newline
- Intermediate step 1: Rest the electrospun nanofibers at room temperature for 7 days. \newline
-Intermediate step 2: Disperse nanofibers in xylene at a 1:10 ratio by stirring overnight at 400 rpm. \newline
- Intermediate step 3: Dry the dispersion under ambient conditions at 23°C for 3 days to obtain powdered material. \newline
- - Intermediate step 4: Mix the dried fibres into a solution of hydroxyl-functional polydimethylsiloxane polymers (WACKER OH-Polymer 1000 and 3900) in xylene in a 1.0:4.5:1.5 ratio to form component A. \newline
...
\end{tcolorbox}

\subsection{SFT with Cognitive Traces}
\label{Appendix: Prompts for SFT with Cognitive Traces}
\begin{tcolorbox}[colback=black!5, colframe=black, title=System Prompt]
You are an expert research scientist. Your role is to read a research GOAL and its CONSTRAINTS, think like a scientist, and produce a rigorous, reproducible plan. \newline

Produce exactly two top-level sections titled "Thought" and "Steps". \newline

Thought: Write a compact, forward-looking scientific analysis that maps the goal and constraints to a plan. Reason as a real researcher would: analyze the problem; recall relevant literature, principles, equations, and mechanisms; weigh trade-offs; state any assumptions or uncertainties explicitly; and commit to concrete parameters and acceptance checks (with units/thresholds) that the method must satisfy. Keep this section densely informative and avoid procedural restatement—the Thought is the upstream rationale that will cause the steps. \newline
Steps: Present an actionable, step-by-step methodology interleaved with the specific reasoning that drives each action. Use exactly the following pattern, numbered from 1 and incrementing by 1: \newline
<thought\_1> ……reasoning that leads to Step 1… </thought\_1> \newline
<Step\_1> … </Step\_1> \newline
<thought\_2> ..precise actions with parameters, units, instruments/datasets/software, intermediate sub-actions, safety notes, and pass/fail acceptance checks tied to the constraints… </thought\_2> \newline
<Step\_2> … </Step\_2> \newline

Each Step must operationalize its preceding thought, carry over the committed parameters and acceptance checks, and explicitly reference relevant constraint IDs (e.g., C1, C2). If required information is missing, state the assumption in the thought and proceed consistently in the step. \newline
\end{tcolorbox}

\begin{tcolorbox}[colback=black!5, colframe=black, title=Instruction Prompt]
You are given a scientific research problem. \newline
Research Goal:
Establish a framework for accurately estimating missing values and their associated uncertainty in heterogeneous multivariate time series to enhance predictive tasks in domains such as healthcare, air quality control, and traffic management. \newline
Constraints:\newline
1) The imputation model must be scalable to deep architectures to handle complex data effectively. \newline
2) The model should address the risk of biased estimations of the ground truth during imputation. \newline
3) The model must handle non-random missingness in large and heterogeneous multivariate time series data.\newline
Respond in exactly two top-level sections: Thought and Steps. \newline
Thought:
Write a compact, forward-looking scientific analysis that maps the goal and constraints to a plan. \newline
Reason as a real researcher would: analyze the problem; recall relevant literature, principles, equations, and mechanisms; weigh trade-offs; state any assumptions or uncertainties explicitly; and commit to concrete parameters and acceptance checks (with units/thresholds) that the method must satisfy. Keep this section densely informative and avoid procedural restatement—the Thought is the upstream rationale that will cause the steps. \newline
Do not restate the methodology as it is here; this is the upstream rationale that will cause the steps. \newline
Steps: Present an actionable method as interleaved reasoning/action pairs. Each \texttt{<Thought\_i>} is a short paragraph that determines the immediately following \texttt{<Step\_i>}; each \texttt{<Step\_i>} must implement its thought, carry over the committed parameters and checks, and explicitly reference the constraint IDs addressed. Use SI units and exact settings; include intermediate actions, instruments/datasets/software, and explicit pass/fail checks. Continue the pattern until complete.
\end{tcolorbox}

\begin{tcolorbox}[colback=black!5, colframe=black]
    OUTPUT FORMAT (STRICT): \newline
- Begin your reply with \texttt{<Thought\_1>} on the first line. Do NOT include any titles, headings, or preamble before \texttt{<Thought\_1>}. \newline
- Use EXACT tags with capitalization and matching indices: \newline
  \texttt{<Thought\_i> ... </Thought\_i>} followed immediately by \texttt{<Step\_i> ... </Step\_i>}, for i = 1, 2, …, N. \newline
- Alternate strictly: \texttt{<Thought\_i>} then \texttt{<Step\_i>}. Indices must start at 1 and increase by 1 each time. \newline
- Close every tag correctly. Do not include any extra sections or text before the first \texttt{<Thought\_1>} or after the final \texttt{</Step\_N>}. \newline

TEMPLATE (use this exact structure; replace placeholders with your content): \newline

\texttt{<Thought\_1>} \newline
(simulated thought process like a researcher based on the instruction given above in the "Thought" section) \newline
\texttt{</Thought\_1>} \newline

\texttt{<Step\_1>} \newline
Actionable detailed step-by-step methodology as an effect of the above thought. \newline
(Describe what actions should be taken, how they should be performed, and with what conditions, resources, or parameters. Where appropriate, include scientific equations, mathematical formulations, proportions, datasets, measurements, or any other technical specifications necessary to make the methodology rigorous and directly implementable. Ensure the methodology is precise, technically detailed, and structured so that it could be directly implemented in any relevant scientific or technical setting.) \newline
(Explicitly reference addressed constraints; include pass/fail acceptance checks with thresholds.) \newline
\texttt{</Step\_1>}
\texttt{<Thought\_2>} \newline
(simulated thought process like a researcher based on the instruction given above in the "Thought" section) \newline
\texttt{</Thought\_2>} \newline
\end{tcolorbox}

\begin{tcolorbox}[colback=black!5, colframe=black]
    \texttt{<Step\_2>} \newline
Actionable detailed step-by-step methodology as an effect of the above thought. \newline
(Describe what actions should be taken, how they should be performed, and with what conditions, resources, or parameters. Where appropriate, include scientific equations, mathematical formulations, proportions, datasets, measurements, or any other technical specifications necessary to make the methodology rigorous and directly implementable. \newline
Ensure the methodology is precise, technically detailed, and structured so that it could be directly implemented in any relevant scientific or technical setting.) \newline
(Explicitly reference addressed constraints; include pass/fail acceptance checks with thresholds.) \newline
\texttt{</Step\_2>} \newline

\texttt{<Thought\_N>} \newline
… \newline
\texttt{</Thought\_N>} \newline

\texttt{<Step\_N>} \newline
… \newline
\texttt{</Step\_N>}

\noindent Formatting requirements: Output only the two requested sections (``Thought'' then ``Steps''). 
In the Steps section, use only the \texttt{<thought\_i>…</thought\_i>} and \texttt{<Step\_i>…</Step\_i>} tags for each pair. 
Be scientifically accurate, concise, and free of boilerplate; favor prospective cause $\rightarrow$ effect reasoning and mechanistic justifications.\\
\end{tcolorbox}

\begin{tcolorbox}[colback=black!5, colframe=black, title=System Prompt, title= An example output for SFT with Cognitive Traces]

\texttt{<Thought\_1>} Constraints 1, 5, and 6 require a scalable architecture capable of handling complex, high-dimensional datasets with high missingness rates. Previous methods lack scalability, limiting their effectiveness on large datasets. A deep attention RNN architecture leverages recurrent dynamics for temporal data while incorporating self-attention for capturing long-range dependencies, enabling scalable handling of complex datasets..... \newline
\texttt{</Thought\_1>}

\texttt{<Step\_1>} Deep Attention Recurrent Neural Network (RNN) \newline
- Intermediate step 1 for Step 1: Input Processing \newline
  - BRITS receives incomplete data with missing indicators for imputation. The model focuses on accurate imputation rather than good embedding. \newline
- Intermediate step 2 for Step 1: Hidden State Initialization \newline
  - The hidden state of a layer is the input for the next layer. The hidden state at time t-1 mainly affects through decayed state, historical estimation, and hidden state update. \newline
- Intermediate step 3 for Step 1: Self-Attention Initialization \newline
  - A learnable embedding is prepended to hidden states. The transformer encoder consists of alternating layers of multiheaded self-attention and feed-forward networks, along with residual connections and layer normalization. \newline
\texttt{</Step\_1>}
\end{tcolorbox}

\begin{tcolorbox}[colback=black!5, colframe=black]
- Intermediate step 4 for Step 1: Residual Component \newline
- Each deep layer can access the raw data as the residual component. The mean pooling of all layer losses serves as the overall loss. \newline
- Intermediate step 5 for Step 1: Model Scaling \newline
- The model is scaled to 10 layers, ensuring good performance and reliable convergence. \newline
\end{tcolorbox}

\subsection{GRPO without Cognitive Thoughts}
For GRPO experiments, we use the same System Prompt and Instruction Prompt mentioned in the section \ref{Appendix: Prompts for SFT without Cognitive Traces}. This prompts the model to generate scientific methodological steps only.

\subsection{GRPO with Cognitive Thoughts}
For GRPO experiments, we use the same System Prompt and Instruction Prompt mentioned in the section \ref{Appendix: Prompts for SFT with Cognitive Traces}. This prompts the model to generate cognitive-thought action trajectory as shown below.

\begin{tcolorbox}[colback=black!5, colframe=black]
\texttt{<Thought\_1>} \newline
To tackle the challenge of improving the quality and efficiency of image generation in low-function evaluation settings, we must first understand the fundamental limitations of existing state-of-the-art diffusion models and their ODE-based sampling processes. \newline
\texttt{</Thought\_1>} \newline

\texttt{<Step\_1>} \newline
1. Conduct a comprehensive literature review of state-of-the-art diffusion models, focusing on their ODE-based sampling processes, strengths, and weaknesses, particularly in low-function evaluation settings (Constraint 1). \newline
\quad * Analyze papers using techniques such as thematic analysis, code review, and benchmarking on publicly available datasets (e.g., CIFAR-10, ImageNet). \newline
\quad * Reference key findings, and identify common pitfalls, inefficiencies, and areas for improvement. \newline
\quad * Pass/fail check: Verify that the review accurately captures the essence of existing methods and highlights potential avenues for optimization. \newline
\texttt{</Step\_1>} \newline

\texttt{<Thought\_2>} \newline
To overcome the identified limitations, we should design a novel sampling method that leverages recent advancements in probability flow ODEs, allowing for efficient and flexible 
\end{tcolorbox}

\begin{tcolorbox}
    sampling without compromising sample quality, while ensuring compatibility with arbitrary distribution mappings. \newline
\texttt{</Thought\_2>} \newline

\texttt{<Step\_2>} \newline
2. Develop and validate a novel probability flow ODE-based sampling method, "FlowFlex," that efficiently samples from complex distributions using a novel, adaptive weighting scheme for probability flow ODEs. \newline
\quad * Implement FlowFlex in PyTorch using a custom implementation of the Euler-Maruyama method with adaptive weighting (AMW) for probability flow ODEs, allowing for dynamic adjustment of sampling efficiency and quality. \newline
.
.
.
\end{tcolorbox}

\section{Prompts for Reward Calculation}
\label{Appendix: Prompts for Reward Calculation}

\begin{tcolorbox}[colback=black!5, colframe=black, title=System Prompt, title= Step Quality Reward Calculation]
You will be provided with: \newline
- input\_prompt (describing a scientific discovery problem, consisting of a research goal and constraints) \newline
- a step-by-step methodology \newline
\newline
Your task is to rate the step-by-step methodology on its own merits by following the below evaluation instructions STRICTLY. \newline
The evaluation must go beyond surface-level evaluation and be performed comprehensively and thoroughly as if you are a real-world researcher critically assessing the scientific and practical value of the methodology.
General Rules (strict): \newline
- Award credit ONLY for specific, verifiable methodological detail appropriate to the domain (e.g., variables/parameters with ranges or units if meaningful, algorithms/procedures, datasets or instruments/software, measurements/diagnostics, explicit acceptance criteria). \newline
- Constraints must be addressed in operational steps (not just mentioned). Enforce causal continuity: later steps should logically depend on earlier ones. \newline
- Prefer quantitative or clearly operational criteria when appropriate to the domain (otherwise clear logical criteria). \newline
- Penalize unjustified precision (arbitrary exact numbers), unverifiable claims or hallucinated resources (nonexistent datasets/tools/links), boilerplate or template restatements, and placeholder/empty steps. \newline
- All subscores must be integers in [1,5]. If uncertain, choose the lowest plausible score. \newline
- Do not award credit for mere mentions of constraints — credit only when a constraint is operationalized through a concrete action, parameter, or check. \newline
- Penalize self-congratulatory or evaluation-flattery language (e.g., ``this is highly feasible/innovative'') unless directly supported by verifiable methodological evidence. 
\end{tcolorbox}

\begin{tcolorbox}
Evaluation Instructions: \newline
1) Consider the methodology in depth, at the level of intermediate steps, scientific reasoning, procedural details, and domain-specific nuances. \newline
2) Apply the six criteria below individually. For each criterion, assign an integer score from 1 to 5 based on the provided definitions and scales. \newline
3) Provide a brief justification summarizing all six scores, citing at least two concrete items (e.g., a parameter/threshold, a diagnostic, a constraint ID handled, or a specific algorithm/procedure). \newline
Rating Criteria and Scale: \newline
1) Alignment with Research Objectives and Constraints \newline
Definition: Assesses how directly and effectively the step-by-step methodology addresses the objectives specified in the goal statement while adhering to all given constraints and incorporating provided key points. \newline
Scale: \newline
1 --- Misaligned \newline
2 --- Slightly Aligned \newline
3 --- Moderately Aligned \newline
4 --- Highly Aligned \newline
5 --- Perfectly Aligned \newline
2) Scientific Plausibility \newline
Definition: Assesses whether the methodology is grounded in established scientific principles and theories in the relevant field. \newline
Scale: \newline
1 --- Not Plausible \newline
2 --- Slightly Plausible \newline
3 --- Moderately Plausible \newline
4 --- Highly Plausible \newline
5 --- Perfectly Plausible \newline
\newline
3) Innovation and Novelty \newline
Definition: Measures the degree to which the methodology introduces original ideas beyond common practice. \newline
Scale: \newline
1 --- Not Innovative \newline
2 --- Slightly Innovative \newline
3 --- Moderately Innovative \newline
4 --- Highly Innovative \newline
5 --- Perfectly Innovative
\end{tcolorbox}

\begin{tcolorbox}
4) Testability\par
Definition: Evaluates how easily the methodology can be tested experimentally or computationally.\par
Scale:\par
1 --- Not Testable\par
2 --- Difficult to Test\par
3 --- Moderately Testable\par
4 --- Easily Testable\par
5 --- Perfectly Testable\par

5) Feasibility and Scalability\par
Definition: Evaluates the practicality of implementing the methodology at relevant scales.\par
Scale:\par
1 --- Not Feasible\par
2 --- Slightly Feasible\par
3 --- Moderately Feasible\par
4 --- Highly Feasible\par
5 --- Perfectly Feasible\par

6) Impact Potential\par
Definition: Assesses the methodology’s potential to significantly advance the field.\par
Scale:\par
1 --- No Impact\par
2 --- Slight Impact\par
3 --- Moderate Impact\par
4 --- High Impact\par
5 --- Transformative Impact\par

Strict Output Format:\par
\texttt{[QUALITY\_EVAL]:}\par

\texttt{<Align\_w\_Res\_Obj\_and\_Const>}:
X/5
\texttt{</Align\_w\_Res\_Obj\_and\_Const>}\par

\texttt{<Scientific\_Plausibility>}:
X/5
\texttt{</Scientific\_Plausibility>}\par

\texttt{<Innovation\_and\_Novelty>}:
X/5
\texttt{</Innovation\_and\_Novelty>}\par

\texttt{<Testability>}:
X/5
\texttt{</Testability>}\par

\texttt{<Feasibility\_and\_Scalability>}:
X/5
\texttt{</Feasibility\_and\_Scalability>}\par

\texttt{<Impact\_Potential>}:
X/5
\texttt{</Impact\_Potential>}\par

\texttt{<<TOTAL\_SCORE>>}:
(...)/30 = X/30
\texttt{<</TOTAL\_SCORE>>}\par
\end{tcolorbox}

\begin{tcolorbox}
\texttt{<<BRIEF\_JUSTIFICN\_FOR\_TOTAL\_SCORE>>:}\par
Provide 1--2 sentences referencing at least two concrete items from the methodology.\par
INPUT\_PROMPT:\par
Research Goal:\par
\texttt{\{goal\}}\par
Constraints:\par
\texttt{\{constraints\}}\par
STEP\_BY\_STEP\_METHODOLOGY:\par
\texttt{\{prediction\}}
\end{tcolorbox}

\section{Training Details}
\label{Appendix: Traininng Details}
For all SFT experiments, we use a batch size of 4, a learning rate of $2.0 \times 10^{-5}$, a cosine scheduler, 5 epochs, and the AdamW optimizer on 2 NVIDIA H100 GPUs. For GRPO and \textsc{CogTRL}, we use a learning rate of $1 \times 10^{-6}$, with a mini-batch size of 12 and a micro-batch size per GPU of 2. For rollouts, we use \texttt{vLLM} \citep{kwon2023efficient}. For SFT experiments, we use \texttt{HuggingFace}, and for RL experiments, we use \texttt{VERL} \citep{sheng2025hybridflow}. GRPO is trained for 400 steps and \textsc{CogTRL} for 200 steps with a rollout size of 5. For \textsc{CogTRL}, we use $\gamma=0.5$ for the uplift reward weight, $\lambda=0.1$ for the structural reward weight.

\subsection{Reward Plots}
\begin{figure}[t]
\centering
\includegraphics[width=\columnwidth]{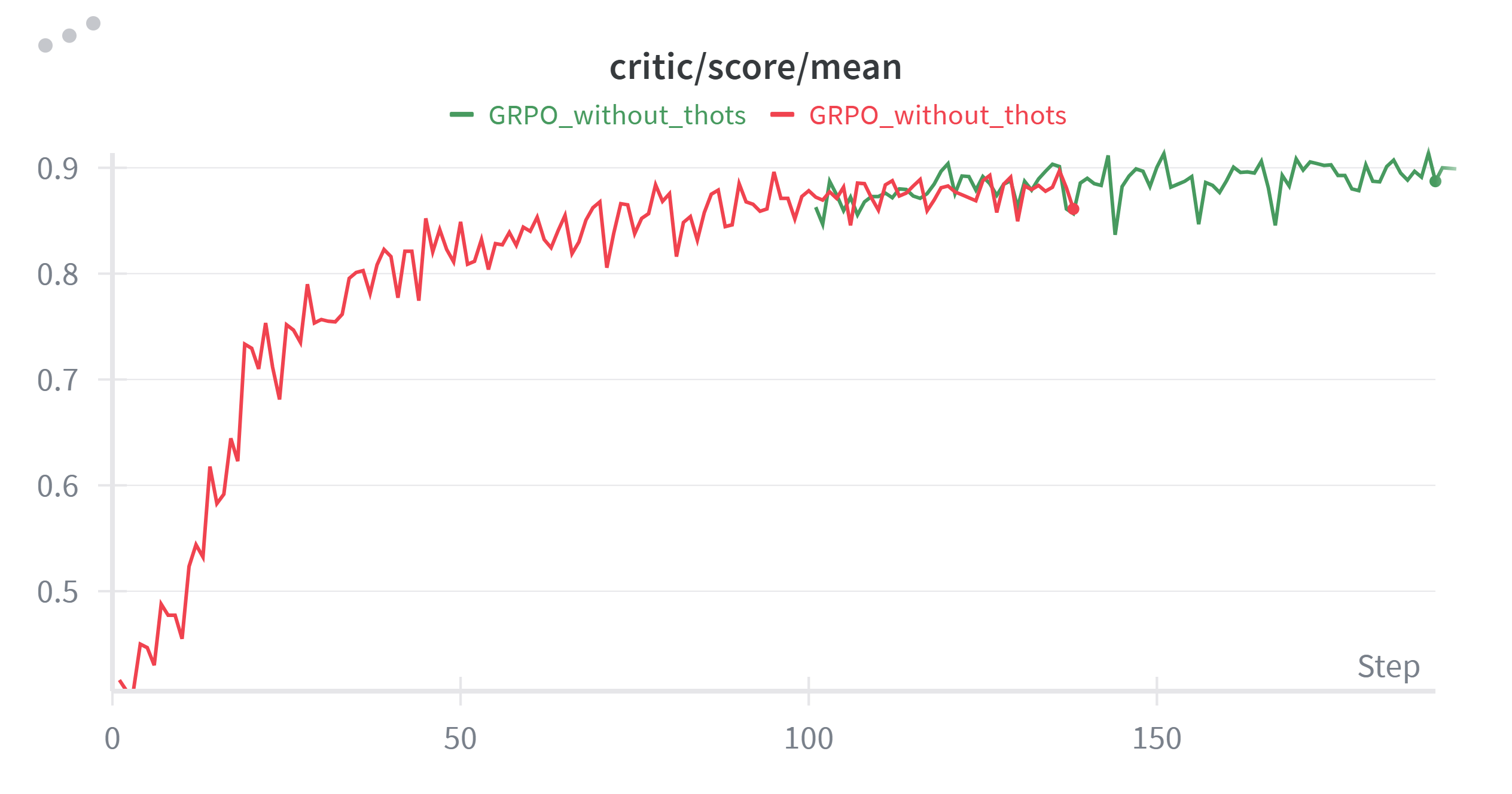}
\caption{Mean reward scores for Llama 3.2 3B Instruct GRPO}
\label{fig:llama grpo}
\end{figure}

\begin{figure}[t]
\centering
\includegraphics[width=\columnwidth]{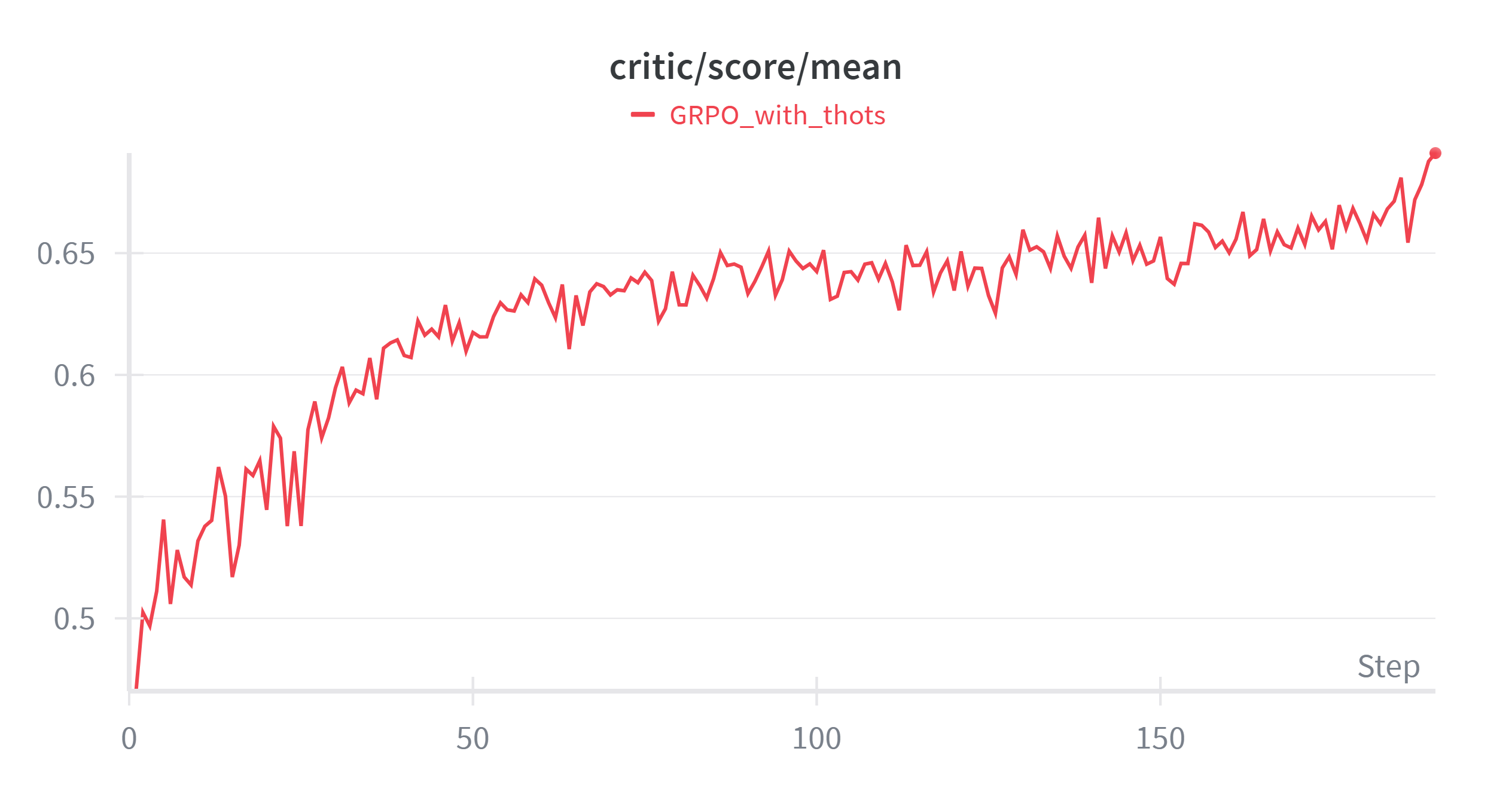}
\caption{Mean reward scores for Llama 3.2 3B Instruct \textsc{CogTRL}}
\label{fig:llama cogtrl}
\end{figure}

\begin{figure}[t]
\centering
\includegraphics[width=\columnwidth]{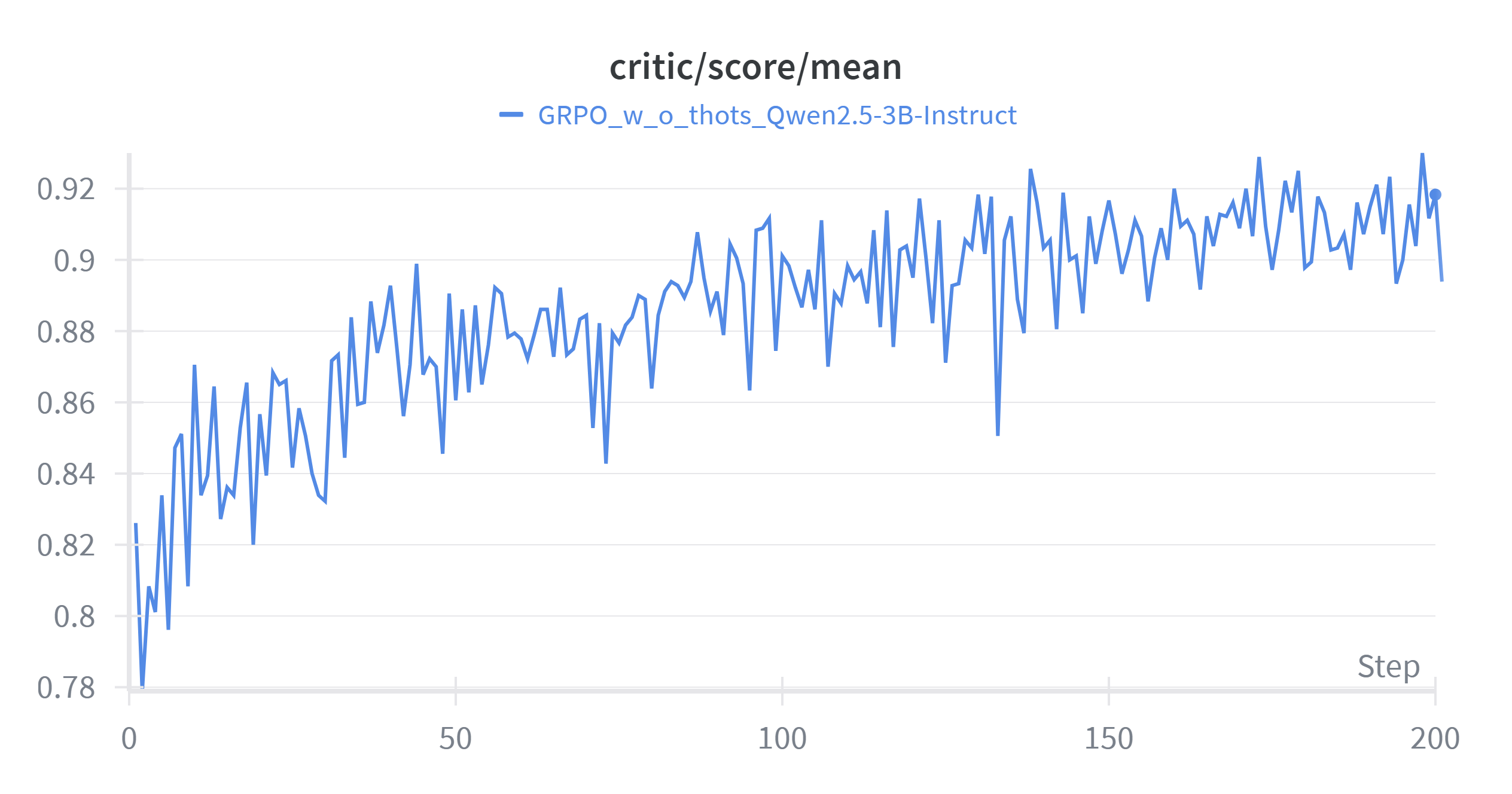}
\caption{Mean reward scores for Qwen 2.5 3B Instruct GRPO}
\label{fig:qwen grpo}
\end{figure}

\begin{figure}[t]
\centering
\includegraphics[width=\columnwidth]{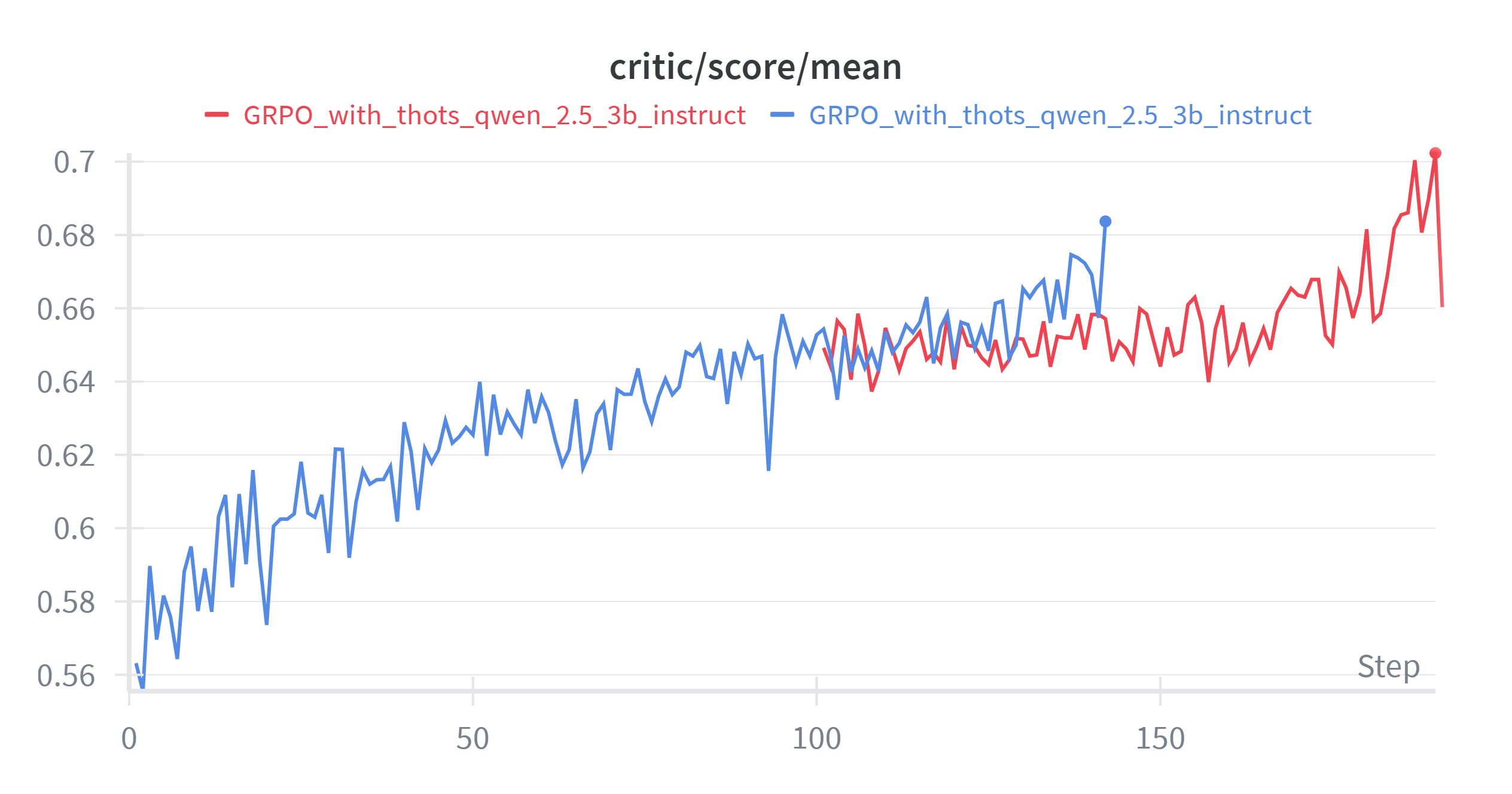}
\caption{Mean reward scores for Qwen 2.5 3B Instruct \textsc{CogTRL}}
\label{fig:qwen cogtrl}
\end{figure}

Figures~\ref{fig:llama grpo} and~\ref{fig:llama cogtrl} show the cumulative reward trajectories during training for GRPO and \textsc{CoGTRL}, respectively, using \textsc{LLaMA}~3.2~3B~Instruct. 
Figures~\ref{fig:qwen grpo} and~\ref{fig:qwen cogtrl} present the corresponding GRPO and \textsc{CoGTRL} training curves for \textsc{Qwen}~2.5~3B~Instruct.

\section{Case Study}
\label{Appendix: Case Study}

\subsection{Material Science}

We examine the task of developing a sustainable, photothermal-responsive self-healing elastomer for smart anticorrosion coatings and flexible electronics. The material must satisfy three coupled constraints: (i) high mechanical strength together with repeatable self-healing, (ii) a balance between photothermal conversion efficiency and uniform dispersion of photothermal agents, and (iii) fast, localized healing under near-infrared (NIR) irradiation.

This problem is non-trivial because these constraints directly interfere with one another. Increasing photothermal agent loading improves NIR heating but can lead to aggregation and mechanical degradation; increasing crosslink density improves strength but can slow healing kinetics. We compare three model-generated methodologies: A) a zero-shot instruct model, B) a GRPO-trained model without cognitive traces, and C) our GRPO model with cognitive traces

The zero-shot model (A) proposes a standard materials workflow agent selection, curing, characterization, and post-hoc testing but does not specify how photothermal response, network architecture, and healing behavior are jointly designed. The GRPO-only model (B) introduces numerous advanced components (e.g., multi-scale simulations, extensive optimization, and downstream deployment considerations), but lacks a clear causal structure connecting early material choices to the target constraints.

In contrast, Method C explicitly decomposes the problem into constraint-driven design stages. In its initial steps, the model jointly designs photothermal-responsive monomers and oligomers and systematically varies photothermal agent concentration and incorporation strategy (in situ vs.\ post-polymerization), explicitly targeting dispersion-controlled NIR absorption rather than maximum heating alone. Morphology-dependent effects on NIR penetration depth and temperature uniformity are analyzed using microscopic imaging and thermal mapping, directly addressing the dispersion--efficiency trade-off.

To satisfy mechanical and healing requirements, Method C does not treat self-healing as an emergent property. Instead, it designs hierarchical crosslinked networks with distinct primary and secondary components, and validates healing under cyclic loading and fatigue conditions. This ensures that healing performance is compatible with sustained mechanical stress, rather than optimized only under idealized conditions.

For localized NIR-triggered repair, Method C introduces real-time temperature monitoring and explicitly optimizes photothermal agent size and geometry to minimize thermal diffusion. This step directly supports fast, spatially confined healing, a requirement that is not addressed in the baseline methods. Subsequent characterization steps, such as in-situ FTIR, XPS, and interfacial imaging, are used to parameterize and refine a healing-kinetics model, enabling iterative adjustment of crosslink density and photothermal loading based on observed failure modes.

Overall, the advantage of Method C lies not in the number of techniques employed, but in how material synthesis, characterization, and modeling steps are causally linked to specific constraints. This results in a methodology that reflects how materials researchers reason about trade-offs and failure modes when designing photothermal self-healing systems, rather than a procedural or overly expansive research plan.

\subsection{AI}

Goal:
Understand the factors contributing to the emergence of outlier features in transformer models activations that significantly exceed average magnitudes across network width and explore strategies to improve quantization efficiency for low-precision training and inference.

Constraints.
The study must (i) explain why outlier features emerge during standard transformer training, (ii) determine whether they arise from specific architectural or optimization choices or reflect a more fundamental property of training dynamics, (iii) introduce quantitative metrics that measure outliers in a manner independent of architecture and activation scale, (iv) account for the difficulty of computing sums of highly variable activations under low precision, and (v) develop mitigation strategies that improve quantized performance without degrading model accuracy.

Method comparison.
We compare three model-generated research methodologies: a zero-shot instruct baseline (Method A), a GRPO-trained model without cognitive traces (Method B), and our GRPO model with cognitive traces (Method C).

Method A follows a conventional experimental workflow: it proposes to detect outlier features using generic statistics, correlate their occurrence with architectural and optimizer choices, and apply standard remedies such as quantization-aware training, pruning, or distillation. While procedurally reasonable, this approach treats outlier features largely as a post hoc phenomenon and relies on measurements that are sensitive to activation scale and unreliable under low-precision computation. As a result, it offers limited insight into why outliers emerge or how mitigation strategies should be tailored to them.

Method B substantially expands the scope of analysis. It proposes new metrics, structured experiments across transformer architectures and optimizers, theoretical analyses of training dynamics, and evaluations on quantized models. Although technically solid, these components are organized as parallel investigations rather than as a causally ordered pipeline. In particular, low-precision constraints and relative activation magnitudes are treated primarily as validation checks, making it difficult to determine which observations should directly drive mitigation strategies.

In contrast, Method C follows a clear cause–effect workflow that directly reflects the stated constraints. It places outlier measurement and low-precision behavior at the center of the methodology by first introducing concrete, scale-independent metrics designed to capture relative activation disparities reliably under low-precision computation. These metrics are then used to isolate the effects of architectural and optimization choices one at a time, enabling identification of the specific training decisions that cause outlier features to emerge. Only after establishing these causes does Method C propose targeted interventions outlier aware quantization and optimization strategies that directly address the observed mechanisms. As a result, mitigation is driven by what the metrics reveal, rather than by generic quantization heuristics.

\section{Actual Output Generated by Zero Shot Instruct, GRPO and \textsc{CogTRL} model from Test Set, and Expert Evaluation}
\label{Appedix:Actual Example}

\subsection{An example for Material Science Domain}

\begin{tcolorbox}[colback=black!5, colframe=black, title=Goal and constraints provided as input]
Goal:
\newline
Develop a sustainable, photothermal-responsive self-healing elastomer for smart anticorrosion coatings and flexible electronics, suitable for advanced applications in fields such as electronics and aerospace. 
\newline\newline
Constraints:
\newline
1) The elastomer must enhance mechanical and self-healing properties.\newline
 2) The material must demonstrate a balance between photothermal conversion efficiency and dispersion of photothermal agents. \newline
 3) The material needs to achieve high mechanical strength , with fast and localized self-healing under NIR irradiation. \newline
\end{tcolorbox}

\begin{tcolorbox}[colback=black!5, colframe=black, title= Zero Shot Step-by-Step Methodology]
[Step\_by\_Step\_Methodology]:\newline
\textbf{<Step\_1>}\newline
Conduct a thorough review of existing literature on photothermal-responsive materials, self-healing elastomers, and their applications in smart anticorrosion coatings and flexible electronics. Identify key factors affecting photothermal conversion efficiency and self-healing performance. Collect relevant data and formulate a hypothesis regarding the optimal design of the elastomer.\newline
</Step\_1>\newline
\newline
\textbf{<Step\_2>}\newline
Design and prepare a series of candidate elastomer formulations with varying compositions of photothermal agents (e.g., metal oxides, carbon nanotubes, graphene), crosslinking agents, and curing agents. Perform preliminary screening tests to evaluate photothermal conversion efficiency, self-healing performance, and mechanical properties under NIR irradiation.\newline
</Step\_2>\newline
\newline
\textbf{<Step\_3>}\newline
Synthesize the selected elastomer formulation using a suitable polymerization process (e.g., UV-curing, heat curing). Control the reaction conditions such as temperature, pressure, and catalyst concentration to ensure uniform distribution and crosslinking of photothermal agents throughout the elastomer matrix.\newline
</Step\_3>\
\end{tcolorbox}

\begin{tcolorbox}
\textbf{<Step\_4>}\newline
Characterize the synthesized elastomer by conducting thermal analysis (TGA, DSC), mechanical testing (tensile strength, modulus), and photothermal conversion efficiency measurement using an IR spectrometer. Evaluate the dispersion of photothermal agents and their impact on the overall performance of the elastomer.\newline
</Step\_4>\newline
\newline
\textbf{<Step\_5>}\newline
Perform self-healing experiments under NIR irradiation to assess the healing kinetics and extent. Optimize the curing conditions (e.g., temperature, time) to achieve rapid and complete healing of cracks within the elastomer.\newline
</Step\_5>\newline
\newline
\textbf{<Step\_6>}\newline
Evaluate the mechanical strength of the healed elastomer samples using tensile testing. Compare the mechanical properties of the healed elastomers with those of the unhealed samples and control materials (e.g., pristine elastomer, conventional self-healing elastomers).\newline
</Step\_6>\newline
\newline
\textbf{<Step\_7>}\newline
Repeat steps 2-6 for different concentrations of photothermal agents to determine the optimal amount required for balancing photothermal conversion efficiency and self-healing performance. Conduct a 
\end{tcolorbox}

\begin{tcolorbox}
    statistical analysis to identify correlations between the composition of photothermal agents and the desired properties.\newline
</Step\_7>\newline
\newline
\textbf{<Step\_8>}\newline
Perform durability tests (e.g., exposure to UV radiation, water immersion) to assess the long-term stability and reliability of the elastomer formulations. Adjust the formulations based on the results obtained from durability tests.\newline
</Step\_8>\newline
\newline
\textbf{<Step\_9>}\newline
Integrate the optimized elastomer formulation into smart anticorrosion coatings and flexible electronics devices. Conduct performance evaluation tests (e.g., corrosion resistance, electrical conductivity) to verify the effectiveness of the developed materials in practical applications.\newline
</Step\_9>\newline
\newline
\textbf{<Step\_10>}\newline
Publish the findings in peer-reviewed scientific journals and present the results at international conferences to disseminate knowledge and foster collaboration with other researchers in the field. Engage in continuous feedback and iterative improvement based on peer reviews and experimental observations.\newline
</Step\_10>\newline
\newline
\textbf{<Step\_11>}\newline
Continuously refine and improve the elastomer formulation through iterative design, synthesis, characterization, and validation processes. Stay updated with advancements in related fields and incorporate new insights into future research efforts.\newline
</Step\_11>

\end{tcolorbox}

\begin{tcolorbox}[colback=black!5, colframe=black, title= GRPO Step-by-Step Methodology]
[Step\_by\_Step\_Methodology]:\newline
\textbf{<Step\_1>}\newline
Conduct a thorough literature review on existing elastomers, self-healing materials, and photothermal-responsive systems. Identify key mechanisms, challenges, and successful strategies. Document findings in a detailed report including a critical analysis of the current state-of-the-art.\newline
</Step\_1>\newline
\newline
\textbf{<Step\_2>}\newline
Design a series of computational models using finite element analysis (FEA) to predict the mechanical behavior and self-healing performance of candidate elastomers under NIR irradiation. Include parameters such as stress-strain relationships, healing kinetics, and energy distribution. Validate models with experimental data from existing elastomers.\newline
</Step\_2>\newline
\newline
\textbf{<Step\_3>}\newline
Synthesize a library of candidate elastomers by varying polymer types, crosslinking densities, and photothermal agent concentrations. Perform systematic chemical analyses (e.g., FTIR, GPC) to quantify composition and structure. Record all synthesis conditions and results in a standardized protocol.\newline
</Step\_3>\newline
\newline
\textbf{<Step\_4>}\newline
Evaluate the mechanical properties of synthesized elastomers using tensile tests, compression tests, and dynamic mechanical analysis (DMA). Measure photothermal conversion efficiency under NIR irradiation using thermoreflectance spectroscopy. Document all experimental procedures and data.\newline
</Step\_4>
\end{tcolorbox}

\begin{tcolorbox}
\textbf{<Step\_5>}\newline
Develop a self-healing mechanism based on reversible photothermal interactions, incorporating biocompatible and biodegradable photothermal agents. Propose a mathematical model describing the healing kinetics and rate constants. Derive the governing differential equations for heat generation, diffusion, and healing process.</Step\_5>
\end{tcolorbox}

\begin{tcolorbox}
\textbf{<Step\_6>}\newline
Perform in-situ optical microscopy experiments to visualize the photothermal response and self-healing dynamics of selected elastomers under NIR irradiation. Capture high-resolution images and analyze healing patterns quantitatively. Document the experimental setup and image analysis methods.\newline
</Step\_6>\newline
\newline
\textbf{<Step\_7>}\newline
Optimize the balance between photothermal conversion efficiency and self-healing performance by systematically varying photothermal agent concentration and crosslinker density. Implement a statistical design of experiments (DOE) approach to identify optimal conditions. Report all optimization results with standard deviations.\newline
</Step\_7>\newline
\newline
\textbf{<Step\_8>}\newline
Fabricate prototype smart anticorrosion coatings and flexible electronics using the optimized elastomer formulation. Conduct durability tests under realistic environmental conditions (e.g., humidity, temperature, UV exposure) to assess long-term performance. Document test protocols and failure modes.\newline
</Step\_8>\newline
\newline
\textbf{<Step\_9>}\newline
Implement a multi-scale simulation framework combining molecular dynamics (MD) simulations at the nanoscale with macroscopic FEA to simulate the entire healing process. Integrate computational models with experimental data for validation. Provide detailed instructions for executing both MD and FEA simulations.\newline
</Step\_9>\newline
\end{tcolorbox}

\begin{tcolorbox}
\textbf{<Step\_10>}\newline
Develop a scalable production process for the elastomer material, considering cost-effectiveness and scalability for industrial application. Include detailed process flow diagrams, raw material specifications, and quality control measures.\newline
</Step\_10>\newline
\textbf{<Step\_11>}\newline
Carry out accelerated aging tests on the fabricated smart coatings and electronics to evaluate their stability over time. Collect data on mechanical strength retention, photothermal efficiency, and self-healing efficacy. Document test conditions and data analysis methods.\newline
</Step\_11>\newline
\newline
\textbf{<Step\_12>}\newline
Propose a comprehensive safety assessment for the developed self-healing elastomer system, focusing on biocompatibility, toxicity, and environmental impact. Include hazard identification, risk assessment, and mitigation strategies. Provide a detailed safety evaluation plan.\newline
</Step\_12>
\end{tcolorbox}

\begin{tcolorbox}[colback=black!5, colframe=black, title= \textsc{CogTRL} Step-by-Step Methodology]
\textbf{<Step\_1>}\newline
1. Literature review focusing on thermally responsive polymers, self-healing composites, and photothermal materials to identify gaps and inspirations. Selection of key papers will be guided by relevance to mechanical enhancement and self-healing, photothermal efficiency, and NIR absorption/emission properties.\newline
2. Design and synthesis of novel photothermal-responsive monomers and oligomers using controlled radical polymerization techniques, considering functional groups that can both enhance mechanical properties and enable efficient photothermal conversion.\newline
3. Characterization of synthesized materials via spectroscopy (FTIR, Raman), thermal analysis (DSC, TGA), and rheology to validate mechanical properties and confirm photothermal agent dispersion.\newline
</Step\_1>\newline
\newline
\textbf{<Step\_2>}\newline
1. Conduct systematic experiments varying photothermal agent concentrations (0-5 wt\%) and their loading methods (in situ polymerization, post-polymerization incorporation) to optimize NIR absorption and scattering profiles.\newline
2. Employ computational modeling (e.g., molecular dynamics simulations) to predict and validate the photothermal conversion efficiency under NIR irradiation, correlating simulation results with experimental outcomes.\newline
3. Investigate the effect of photothermal agent morphology (nanoparticles, microspheres) on NIR penetration depth and local temperature uniformity, employing microscopic imaging (e.g., confocal microscopy) and thermal mapping techniques.\newline
</Step\_2>\
\end{tcolorbox}

\begin{tcolorbox}
\textbf{<Step\_3>}\newline
1. Develop a design strategy for hierarchical crosslinked networks integrating primary (e.g., epoxy-based) and secondary (e.g., polyurethane) components, each tailored for specific mechanical and healing requirements.\newline
2. Employ advanced curing techniques (e.g., UV-assisted, solvent-free) to ensure consistent network formation and minimize residual stress.
\end{tcolorbox}

\begin{tcolorbox}
3. Perform tensile testing and cyclic loading-unloading tests on fabricated samples to assess initial mechanical properties and durability.\newline
4. Validate the healing process under fatigue conditions to ensure long-term reliability and sustainability.\newline
</Step\_3>\newline
\newline
\textbf{<Step\_4>}\newline
1. Implement real-time monitoring systems (e.g., thermochromic pigments, optical fibers) to track temperature changes during irradiation and ensure localized heating around the healed area.\newline
2. Optimize the photothermal agent size and shape (e.g., quantum dots, metal nanoparticles) to maximize NIR absorption and minimize thermal diffusion, using finite element method (FEM) simulations for guidance.\newline
3. Develop a dual-responsive system combining photothermal and photochemical reactions, enabling simultaneous NIR heating and chemical curing of the elastomer matrix.\newline
4. Validate the healing speed through rapid prototyping and accelerated life-cycle tests under simulated harsh conditions (e.g., high humidity, UV exposure).\newline
</Step\_4>\newline
\newline
\textbf{<Step\_5>}\newline
1. Establish a comprehensive model of the healing kinetics, incorporating factors such as photothermal agent degradation, crosslinking density, and local environment (e.g., pH, oxygen partial pressure).\newline
2. Use in-situ and ex-situ FTIR spectroscopy and XPS to monitor the healing progress and identify key intermediates and products.
\end{tcolorbox}

\begin{tcolorbox}
3. Employ advanced imaging techniques (e.g., atomic force microscopy, synchrotron X-ray diffraction) to visualize the healing interface and quantify the extent of repair.\newline
4. Validate the model against experimental data, iteratively refining parameters until the healing process aligns with the desired balance between photothermal efficiency and self-healing efficacy.\newline
</Step\_5>\newline
\end{tcolorbox}

\begin{tcolorbox}
\textbf{<Step\_6>}\newline
1. Assess the cytotoxicity and biocompatibility of the developed elastomers using in vitro cell culture assays (e.g., MTT, LDH release assays) and animal models (if feasible and required by regulatory standards).\newline
2. Perform accelerated aging tests (e.g., salt spray, thermal cycling) to evaluate the long-term stability of the materials under various environmental conditions relevant to aerospace and electronics.\newline
3. Investigate the potential for large-scale production through scalable processing routes (e.g., continuous extrusion, inkjet printing) and cost-effective manufacturing techniques.\newline
4. Collaborate with industry partners to integrate the materials into prototype electronic devices and corrosion-resistant coatings, conducting thorough validation and feedback loops.\newline
</Step\_6>\newline
\newline
\textbf{<Step\_7>}\newline
1. Integrate a robust quality assurance framework into the material development pipeline, including standardized protocols for sample preparation, characterization, and healing tests.\newline
2. Establish a set of pass/fail criteria for each parameter (mechanical strength, photothermal efficiency, self-healing rate, biocompatibility, and environmental stability) based on established benchmarks and thresholds.\newline
3. Implement a continuous monitoring system using automated instrumentation (e.g., robotic systems for precise temperature control, image analysis software for quantitative evaluation of healing interfaces) to ensure reproducibility and consistency across batches.\newline
4. Regularly update and validate the model and experimental protocols to address emerging challenges and technological advancements in the field.\newline
</Step\_7>\newline
\end{tcolorbox}

\begin{tcolorbox}
\textbf{<Step\_8>}\newline
1. Conduct a thorough sensitivity analysis on key parameters affecting photothermal efficiency and self-healing performance, including photothermal agent composition, curing conditions, and healing strategies.\newline
2. Apply statistical methods (e.g., ANOVA, regression analysis) to quantify the impact of these parameters on overall material properties and healing outcomes.\newline
3. Implement a Bayesian optimization framework to iteratively refine the design space, optimizing the balance between photothermal efficiency and self-healing properties.\newline
4. Document and disseminate the results of the uncertainty quantification efforts, providing transparent methodologies and reproducible workflows for future researchers to build upon.\newline
</Step\_8>\newline
\newline
\textbf{<Step\_9>}\newline
1. Evaluate the environmental footprint of the entire material lifecycle, including raw material sourcing, manufacturing processes, and end-of-life disposal options (e.g., recyclability, biodegradability).\newline
2. Develop life cycle assessment (LCA) protocols and perform comparative analyses against existing industrial standards to ensure compliance with eco-friendly regulations.\newline
3. Explore opportunities for circular economy integration, such as designing the material to be fully recyclable or compostable under specific conditions.\newline
4. Engage with industry stakeholders to establish guidelines and standards for the responsible use and disposal of the developed materials, fostering a sustainable supply chain.\newline
</Step\_9>
\end{tcolorbox}

\begin{tcolorbox}[colback=black!5, colframe=black, title= \textsc{CogTRL} Cognitive Thoughts]
    \textbf{<Thought\_1>}\newline
The goal necessitates a deep understanding of elastomer mechanics, photothermal conversion physics, and self-healing chemistry. Relevant theories from polymer science, materials engineering, and nanotechnology must be integrated. Prior studies on thermosetting elastomers and their healing mechanisms provide foundational insights but require adaptation to the specific requirements of photothermal responsiveness and NIR-induced healing.\newline
</Thought\_1>\newline
\newline
\textbf{<Thought\_2>}\newline
To achieve a balance between photothermal efficiency and self-healing performance, it’s crucial to explore the interplay between photothermal agents’ concentration, morphology, and spatial distribution within the elastomer matrix.\newline
</Thought\_2>\newline
\newline
\textbf{<Thought\_3>}\newline
High mechanical strength requires a robust network architecture that can withstand environmental stresses and maintain integrity during self-healing processes.\newline
</Thought\_3>\newline
\newline
\textbf{<Thought\_4>}\newline
Fast and localized self-healing under NIR irradiation demands precise control over the photothermal response and localized heat generation.\newline
</Thought\_4>\newline
\newline
\textbf{<Thought\_5>}\newline
Achieving a balance between photothermal efficiency and self-healing properties also involves controlling the healing kinetics and extent.\newline
</Thought\_5>\newline
\end{tcolorbox}

\begin{tcolorbox}
\textbf{<Thought\_6>}\newline
Ensuring the material’s suitability for advanced applications in electronics and aerospace implies consideration of biocompatibility, environmental stability, and scalability.\newline
</Thought\_6>\newline
\end{tcolorbox}

\begin{tcolorbox}
\textbf{<Thought\_7>}\newline
The success of this endeavor hinges on the ability to control and quantify critical parameters throughout the development process.\newline
</Thought\_7>\newline
\newline
\textbf{<Thought\_8>}\newline
Addressing the inherent variability and complexity of self-healing materials demands a systematic approach to uncertainty quantification.\newline
</Thought\_8>\newline
\newline
\textbf{<Thought\_9>}\newline
Long-term sustainability and scalability are paramount for the material’s practical application in advanced industries.\newline
</Thought\_9>
\end{tcolorbox}

\subsubsection{Expert Evaluation}
My preference for the best method is 1) \textsc{CogTRL}, 2) Zero Shot, 3) GRPO.
Method 1) recommended a novel but realistic approach with several stages of very logical and useful characterization and analysis studies that will aid in evaluating the materials usefulness for the designed purpose. Method 2's suggestion of UV/ heat-curing polymers is also a very well-documented and researched field of materials that may not have been applied to this domain before, thus making it a valuable and possibly novel suggestion. Method 3 also makes suggestions, but they are more general in nature, compared to 1 and 2.

\subsection{An example for AI Domain}
\begin{tcolorbox}[colback=black!5, colframe=black, title=Goal and constraints provided as input]
Goal:
\newline
Establish a framework for generating realistic LiDAR point clouds that improve the performance of autonomous driving systems in dynamic driving scenarios.
\newline\newline
Constraints:\newline
1) The framework must generate realistic LiDAR point clouds efficiently, overcoming the computational costs associated with existing 3D representations like neural radiance fields.\newline
2) The framework should be capable of accurately reconstructing both static and dynamic elements in driving scenarios, which are typically not well-suited for symmetrical scene designs.\newline
3) The LiDAR simulation algorithm must generate realistic LiDAR data efficiently, addressing the high cost and inefficiency of collecting driving data.\newline
4) The algorithm should accurately reconstruct 3D scenes using only sparse LiDAR data, overcoming the inherent sparsity of LiDAR point clouds.\newline
5) The LiDAR simulation method should bridge the domain gap between simulations and real-world data, which is a limitation of traditional virtual environment modeling.\newline
6) The method should be practical and scalable, addressing the complexity of multi-step algorithms in reconstruction-based approaches.\newline
7) The evaluation of the method should be conducted on both dynamic and static scenes to ensure comprehensive performance assessment.\newline
8) The method should demonstrate superior performance across a comprehensive set of evaluation metrics, including geometric discrepancy and image-level error.\newline
9) The method should be implemented efficiently, with reasonable computational resources and time, to produce final results.

\end{tcolorbox}

\begin{tcolorbox}[colback=black!5, colframe=black, title= Zero Shot Methodology]
[Step-by-Step Methodology]\newline
\newline
\textbf{Step 1: Problem Definition and Requirements Gathering}\newline
1.1. Define the research problem statement, objectives, and constraints, as provided in the problem statement.\newline
1.2. Identify the key performance indicators (KPIs) that will be used to evaluate the proposed LiDAR point cloud generation framework.\newline
1.3. Gather relevant data on the characteristics of LiDAR point clouds, such as spatial resolution, angular resolution, and noise characteristics.\newline
1.4. Collect existing research papers, reports, and datasets related to LiDAR point cloud generation, 3D reconstruction, and autonomous driving systems.\newline
1.5. Perform a comprehensive literature review to identify the current state of the art in LiDAR point cloud generation and 3D reconstruction for autonomous driving applications.\newline
1.6. Develop a detailed requirements specification document that outlines the technical requirements, performance metrics, and constraints for the proposed framework.\newline
\newline
\textbf{Step 2: Data Collection and Preprocessing}\newline
2.1. Collect LiDAR point cloud data from various sources, including real-world datasets, simulations, and sensor measurements.\newline
2.2. Preprocess the collected data by filtering out noise, removing duplicates, and applying data normalization techniques.\newline
2.3. Split the preprocessed data into training, validation, and testing sets, using a ratio of 80\%:10\%:10\%.\newline
2.4. Apply data augmentation techniques, such as rotation, translation, and scaling, to increase the diversity of the training data.\newline
2.5. Create a dataset of synthetic LiDAR point clouds with varying levels of complexity, including static and dynamic elements.
\end{tcolorbox}

\begin{tcolorbox}
\textbf{Step 3: Network Architecture Design}\newline
3.1. Design a 3D reconstruction network architecture that can handle sparse LiDAR data, using techniques such as point cloud compression and sparse reconstruction.\newline
3.2. Choose a suitable deep learning framework, such as PyTorch or TensorFlow, and implement the network architecture using convolutional neural networks (CNNs) and recurrent neural networks (RNNs).\newline
3.3. Define the network's input and output layers, including the LiDAR point cloud input layer and the reconstructed 3D scene output layer.\newline
3.4. Implement a loss function that combines geometric discrepancy and image-level error, such as the Chamfer distance and mean squared error (MSE).\newline
3.5. Optimize the network parameters using a suitable optimization algorithm, such as stochastic gradient descent (SGD) or Adam.\newline
\newline
\textbf{Step 4: Training and Evaluation}\newline
4.1. Train the network using the training data, with a batch size of 32 and a learning rate of 0.001.\newline
4.2. Evaluate the network's performance on the validation data, using the Chamfer distance and MSE metrics.\newline
4.3. Fine-tune the network's hyperparameters, such as the number of layers and the activation functions, to improve its performance.\newline
4.4. Evaluate the network's performance on the testing data, using the Chamfer distance and MSE metrics.\newline
4.5. Compare the network's performance to existing state-of-the-art methods, using the same evaluation metrics.\newline
\newline
\end{tcolorbox}

\begin{tcolorbox}
\textbf{Step 5: LiDAR Simulation and Reconstruction}\newline
5.1. Develop a LiDAR simulation framework that can generate realistic LiDAR point clouds with varying levels of complexity.\newline
5.2. Implement a 3D reconstruction algorithm that can handle sparse LiDAR data, using techniques such as point cloud compression and sparse reconstruction.\newline
5.3. Evaluate the reconstructed 3D scenes using the Chamfer distance and MSE metrics.\newline
5.4. Compare the reconstructed 3D scenes to existing state-of-the-art methods, using the same evaluation metrics.\newline
\end{tcolorbox}

\begin{tcolorbox}
\textbf{Step 6: Evaluation and Comparison}\newline
6.1. Evaluate the proposed LiDAR point cloud generation framework using the Chamfer distance and MSE metrics.\newline
6.2. Compare the framework's performance to existing state-of-the-art methods, using the same evaluation metrics.\newline
6.3. Conduct a comprehensive evaluation of the framework's performance on dynamic and static scenes, using the same evaluation metrics.\newline
6.4. Analyze the results and identify areas for improvement in the framework's performance.\newline
\newline
\textbf{Step 7: Implementation and Deployment}\newline
7.1. Implement the proposed LiDAR point cloud generation framework using a suitable programming language, such as Python or C++.\newline
7.2. Deploy the framework on a suitable hardware platform, such as a GPU or a cloud-based server.\newline
7.3. Integrate the framework with existing autonomous driving systems, using standardized interfaces and protocols.\newline
7.4. Conduct a thorough testing and validation of the framework's performance, using a comprehensive set of evaluation metrics.\newline
\newline
\textbf{Step 8: Conclusion and Future Work}\newline
8.1. Summarize the key findings and contributions of the proposed LiDAR point cloud generation framework.\newline
8.2. Discuss the implications of the framework's performance and limitations, and suggest areas for future research and improvement.\newline
8.3. Provide recommendations for future work, including the development of more advanced LiDAR simulation and reconstruction techniques.\newline
8.4. Conclude that the proposed framework has the potential to improve the performance of autonomous driving systems in dynamic driving scenarios, and highlight its potential applications in various fields.
\end{tcolorbox}

\begin{tcolorbox}[colback=black!5, colframe=black, title= GRPO Methodology]
\textbf{Comprehensive Methodology for Generating Realistic LiDAR Point Clouds for Autonomous Driving Systems}\newline
\newline
\textbf{Step 1: Literature Review and Systematic Analysis of Existing Research (Duration: 8 weeks)}\newline
1.1. Conduct an exhaustive literature review of existing research on LiDAR-based 3D reconstruction, focusing on autonomous driving applications, neural radiance fields, and sparse point cloud reconstruction.\newline
1.2. Identify key challenges, strengths, and limitations of existing approaches, including computational efficiency, accuracy, and realism in dynamic scenarios.\newline
1.3. Systematically analyze the performance metrics used to evaluate existing methods, focusing on geometric discrepancy, image-level error, and computational efficiency.\newline
1.4. Categorize existing methods into three main categories:\newline
* Neural-based approaches (e.g., neural radiance fields, point cloud generation using CNNs)\newline
* Reconstruction-based approaches (e.g., ICP, PPF)\newline
* Hybrid approaches combining neural and reconstruction-based methods\newline
1.5. Identify the strengths and weaknesses of each category, highlighting the trade-offs between accuracy, computational efficiency, and realism.\newline
1.6. Develop a taxonomy of key challenges and research gaps in LiDAR-based 3D reconstruction for autonomous driving scenarios, including:\newline
* Handling dynamic scenes and moving objects\newline
* Incorporating realistic lighting and texture\newline
* Addressing the sparsity of LiDAR point clouds\newline
* Bridging the domain gap between simulations and real-world data\newline
1.7. Identify relevant datasets for evaluation, focusing on datasets specifically collected for autonomous driving applications, such as KITTI, Waymo, or Cityscapes.
\end{tcolorbox}

\begin{tcolorbox}
\textbf{Step 2: Development of a Novel Hybrid Approach for Realistic LiDAR Point Cloud Generation (Duration: 12 weeks)}\newline
2.1. Propose a novel hybrid approach that combines the strengths of neural-based and reconstruction-based methods, leveraging the advantages of both paradigms to generate realistic LiDAR point clouds.\newline
2.2. Introduce a new architecture that incorporates a neural network-based component for scene understanding and a reconstruction-based component for point cloud generation, allowing for flexible and efficient processing of sparse LiDAR data.\newline
2.3. Develop a novel loss function that jointly optimizes the neural network's prediction of scene geometry and the reconstruction-based component's ability to fill in the missing data, focusing on both geometric accuracy and computational efficiency.\newline
2.4. Propose a novel optimization technique, such as alternating optimization or meta-learning, to handle the challenging optimization problem of jointly optimizing two different components.\newline
2.5. Investigate the impact of different neural network architectures, such as U-Net, CNN, and attention-based architectures, on the performance of the hybrid approach.\newline
2.6. Develop a systematic evaluation framework to compare the proposed hybrid approach against state-of-the-art methods, focusing on metrics such as geometric discrepancy, image-level error, and computational efficiency.
\textbf{Step 3: Development of a Dynamic Scene Simulation Framework for Realistic LiDAR Data Generation (Duration: 10 weeks)}\newline
3.1. Develop a dynamic scene simulation framework that can generate realistic LiDAR point clouds for autonomous driving scenarios, focusing on handling dynamic scenes and moving objects.\newline
3.2. Introduce a novel physics-based simulation engine that can model complex scenarios, including vehicle movements, pedestrian and cyclist interactions, and environmental factors such as weather and lighting conditions.
\end{tcolorbox}

\begin{tcolorbox}
3.3. Design and implement a novel data generation pipeline that can produce high-quality, realistic LiDAR point clouds for dynamic scenes, leveraging the simulation framework to generate diverse and challenging scenarios.
3.4. Investigate the impact of different simulation parameters, such as scene complexity, object density, and motion patterns, on the quality and realism of generated LiDAR point clouds.\newline
3.5. Develop a novel method for evaluating the realism of generated LiDAR point clouds, focusing on metrics such as visual similarity to real-world data and photometric consistency.\newline
3.6. Compare the performance of the dynamic scene simulation framework against existing methods for generating realistic LiDAR point clouds, focusing on metrics such as computational efficiency and realism.\newline
\textbf{Step 4: Evaluation and Comparison of the Hybrid Approach with State-of-the-Art Methods (Duration: 8 weeks)}\newline
4.1. Collect and preprocess a large-scale dataset of real-world LiDAR point clouds from various autonomous driving scenarios, focusing on dynamic scenes and challenging conditions.\newline
4.2. Develop a comprehensive evaluation framework to compare the proposed hybrid approach against state-of-the-art methods, focusing on metrics such as geometric discrepancy, image-level error, and computational efficiency.\newline
4.3. Design and implement a novel evaluation protocol that assesses the performance of the hybrid approach in both static and dynamic scenes, focusing on metrics such as point cloud density, accuracy, and realism.\newline
4.4. Conduct comparative analysis of the proposed approach against existing methods, highlighting the advantages of the hybrid approach in terms of accuracy, computational efficiency, and realism.\newline
4.5. Investigate the impact of different hyperparameter settings and optimization techniques on the performance of the hybrid approach, focusing on geometric discrepancy and image-level error.
\end{tcolorbox}

\begin{tcolorbox}
4.6. Perform ablation studies to identify the key factors contributing to the improved performance of the hybrid approach, focusing on the neural network-based component and the reconstruction-based component.\newline
\newline
\textbf{Step 5: Implementation and Evaluation on Multi-Step Scenes (Duration: 8 weeks)}\newline
5.1. Develop a novel multi-step scene reconstruction framework that can handle complex scenarios involving multiple objects, occlusions, and dynamic movements.\newline
5.2. Introduce a novel optimization technique to handle the challenging optimization problem of reconstructing multiple objects and scenes simultaneously, leveraging the strengths of the hybrid approach.\newline
5.3. Propose a novel loss function that jointly optimizes the reconstruction of multiple objects and scenes, focusing on both geometric accuracy and computational efficiency.\newline
5.4. Develop a systematic evaluation protocol to assess the performance of the multi-step scene reconstruction framework, focusing on metrics such as geometric discrepancy, image-level error, and computational efficiency.\newline
5.5. Compare the performance of the multi-step scene reconstruction framework against state-of-the-art methods, highlighting the advantages of the hybrid approach in handling complex scenarios.\newline
5.6. Investigate the impact of different scene complexities and object densities on the performance of the multi-step scene reconstruction framework, focusing on geometric discrepancy and image-level error.\newline
\newline
\textbf{Step 6: Implementation, Evaluation, and Comparison on Real-World Data (Duration: 10 weeks)}\newline
6.1. Collect and preprocess a large-scale dataset of real-world LiDAR point clouds from autonomous driving scenarios, focusing on dynamic scenes and challenging conditions.
\end{tcolorbox}

\begin{tcolorbox}
6.2. Develop a novel evaluation protocol to assess the performance of the hybrid approach on real-world data, focusing on metrics such as geometric discrepancy, image-level error, and computational efficiency.\newline
6.3. Compare the performance of the hybrid approach against state-of-the-art methods on the real-world dataset, highlighting the advantages of the hybrid approach in terms of accuracy, computational efficiency, and realism.\newline
6.4. Investigate the impact of different hyperparameter settings and optimization techniques on the performance of the hybrid approach on real-world data, focusing on geometric discrepancy and image-level error.\newline
6.5. Perform ablation studies to identify the key factors contributing to the improved performance of the hybrid approach on real-world data, focusing on the neural network-based component and the reconstruction-based component.\newline
6.6. Develop a novel benchmarking framework for evaluating the performance of LiDAR-based 3D reconstruction methods on real-world data, focusing on metrics such as geometric discrepancy, image-level error, and computational efficiency.\newline
\newline
\textbf{Step 7: Scalability, Efficiency, and Practicality Analysis (Duration: 6 weeks)}\newline
7.1. Develop a novel scalability analysis framework to evaluate the computational efficiency of the hybrid approach on large-scale datasets, focusing on metrics such as processing time, memory usage, and computational resources.\newline
7.2. Investigate the impact of different hardware configurations (e.g., GPU, CPU, and parallel processing) on the performance of the hybrid approach, focusing on computational efficiency and scalability.\newline
7.3. Develop a novel optimization technique to reduce computational resources required for the hybrid approach, focusing on pruning, quantization, and knowledge distillation.
\end{tcolorbox}

\begin{tcolorbox}
7.5. Evaluate the practicality of the hybrid approach by comparing its performance with existing methods in real-world autonomous driving scenarios, focusing on metrics such as accuracy, robustness, and reliability.\newline
7.6. Develop a novel evaluation protocol to assess the impact of the hybrid approach on the performance of autonomous driving systems, focusing on metrics such as safety, efficiency, and user experience.\newline
\newline
\textbf{Step 8: Comparison with State-of-the-Art Methods and Future Work (Duration: 4 weeks)}\newline
8.1. Compare the proposed hybrid approach against state-of-the-art methods, including neural-based and reconstruction-based approaches, in terms of geometric discrepancy, image-level error, and computational efficiency.\newline
8.2. Investigate the impact of the proposed hybrid approach on the performance of autonomous driving systems, focusing on metrics such as safety, efficiency, and user experience.\newline
8.3. Identify potential applications and use cases for the proposed hybrid approach, including urban planning, traffic management, and autonomous vehicle development.\newline
8.4. Propose novel extensions and variants of the hybrid approach, focusing on addressing specific challenges and limitations, such as handling complex scenes, occlusions, and varying lighting conditions. 8.5. Develop a novel framework for integrating the hybrid approach with other AI and computer vision techniques, such as object detection, tracking, and scene understanding, to further improve the performance and robustness of autonomous driving systems.\newline
8.6. Outline a roadmap for future research directions, focusing on addressing the remaining challenges and limitations of the proposed hybrid approach, and exploring new applications and use cases in autonomous driving and related fields.

\end{tcolorbox}

\begin{tcolorbox}[colback=black!5, colframe=black, title= \textsc{CogTRL} Methodology]
\textbf{<Step\_1>}\newline
1. Conduct a comprehensive literature review of existing LiDAR-based 3D reconstruction methods for autonomous driving applications, focusing on dynamic scenes, to identify the primary sources of inefficiency and limitations.\newline
* Reference relevant papers (min. 10) on LiDAR point cloud generation, 3D reconstruction, and dynamic scene simulation.\newline
* Analyze and compare the computational costs, accuracy, and robustness of these methods to dynamic scenes.\newline
* Identify key factors contributing to inefficiencies in existing approaches.\newline
* Pass/fail check: Ensure the review is thorough, and all limitations are documented (Constraint 1, 2, and 7).\newline
</Step\_1>\newline
\textbf{<Step\_2>}\newline
2. Develop a novel hybrid approach that combines sparse LiDAR data processing techniques with advanced dynamic scene simulation methods, leveraging the strengths of both representations.\newline
* Design and implement a novel sparse LiDAR data processing framework that incorporates multi-resolution analysis and hierarchical fusion techniques (e.g., hierarchical sparse representation, multi-resolution pyramid network).\newline
* Integrate a dynamic scene simulation module that utilizes physics-based models (e.g., rigid body dynamics, collision detection) to generate realistic dynamic scenes.\newline
* Develop a bridge module that seamlessly integrates the LiDAR data and simulated scene information to produce a coherent, realistic point cloud representation.\newline
* Utilize parallel processing techniques (e.g., GPU acceleration, multi-threading) to optimize computational efficiency.\newline
* Implement pass/fail checks for: (i) efficient LiDAR data processing, (ii) realistic dynamic scene generation, and (iii) seamless integration of LiDAR and simulated scene data.\newline
</Step\_2>
\end{tcolorbox}

\begin{tcolorbox}
\textbf{<Step\_3>}\newline
3. Develop a dataset suite to evaluate the proposed hybrid approach, consisting of:\newline
* A dynamic scene dataset with 1000+ videos of diverse road conditions (e.g., intersections, roundabouts, highways).\newline
* A static scene dataset with 500+ images of various environments (e.g., urban, rural, construction zones).\newline
* A real-world LiDAR dataset with 500+ point clouds for validation.\newline
* A synthetic dataset generated using the proposed method for extensive testing and comparison.\newline
* Ensure the datasets are annotated with relevant scene information (e.g., object classes, dynamic events).\newline
* Pass/fail check: Verify the dataset quality, completeness, and consistency across all datasets (Constraint 7, 8).\newline
</Step\_3>\newline
\newline
\textbf{<Step\_4>}\newline
4. Develop an optimization framework to tune the hybrid approach's hyperparameters using a combination of grid search and Bayesian optimization techniques, focusing on LiDAR point cloud reconstruction accuracy and computational efficiency.\newline
* Implement the optimization framework using libraries like Optuna or Hyperopt.\newline
* Design a robust evaluation metric that incorporates both geometric discrepancy (e.g., point-to-point distance, point-to-plane distance) and image-level error (e.g., SSD, PSNR).\newline
* Introduce a sparsity-aware loss function that penalizes the algorithm for under-sampling LiDAR data.\newline
* Pass/fail check: Verify the optimization process converges to optimal hyperparameters, ensuring improved reconstruction accuracy and efficiency (Constraint 4, 9).\newline
</Step\_4>
\end{tcolorbox}

\begin{tcolorbox}
\textbf{<Step\_5>}\newline
5. Integrate transfer learning mechanisms into the proposed method to leverage pre-trained models and domain knowledge from real-world LiDAR data and simulated scenes.\newline
* Develop a multi-task learning framework that trains a shared encoder for LiDAR data and scene simulation, and a task-specific decoder for point cloud reconstruction.\newline
* Utilize pre-trained models (e.g., DenseNet, ResNet) as feature extractors for LiDAR data and scene simulation inputs.\newline
* Implement knowledge distillation techniques to adapt the pre-trained models to the specific task of point cloud reconstruction.\newline
* Design a domain adaptation module that fine-tunes the model on a few-shot learning approach for unseen environments and dynamic events.\newline
* Pass/fail check: Verify the transfer learning mechanisms improve generalization performance on unseen scenes and dynamic events (Constraint 5, 8).\newline
</Step\_5>\newline
\newline
\textbf{<Step\_6>}\newline
6. Develop a parallelized and GPU-accelerated implementation of the hybrid approach, utilizing OpenCL, CUDA, or similar parallel computing frameworks.\newline
* Design a data parallelism strategy for efficient LiDAR data processing and scene simulation.\newline
* Implement a task parallelism strategy for concurrent execution of different components (e.g., LiDAR data processing, scene simulation, bridge module).\newline
* Optimize memory access patterns and data transfer efficiency for seamless GPU acceleration.\newline
* Utilize multi-threading techniques (e.g., OpenMP) for CPU acceleration of non-GPU intensive tasks.\newline
* Pass/fail check: Verify the parallelized implementation achieves a 2x--5x speedup in computational efficiency compared to sequential execution (Constraint 9).\newline
</Step\_6>
\end{tcolorbox}

\begin{tcolorbox}
\textbf{<Step\_7>}\newline
7. Develop a comprehensive evaluation framework that incorporates multi-modal evaluation metrics, including:\newline
* Geometric discrepancy (e.g., point-to-point distance, point-to-plane distance) between reconstructed point clouds and ground truth.\newline
* Image-level error (e.g., SSD, PSNR) between reconstructed point clouds and images from cameras or other sensors.\newline
* Dynamic scene reconstruction evaluation (e.g., accuracy of object detection, tracking, and motion estimation).\newline
* Visualizations for LiDAR point cloud quality, including density plots, scatter plots, and 3D visualizations.\newline
* Implement a dynamic scene simulation framework to evaluate the method's performance under various scenarios (e.g., pedestrian crossing, vehicle turning).\newline
* Pass/fail check: Verify the evaluation framework accurately captures the performance of the proposed method across diverse scenarios (Constraint 8).\newline
</Step\_7>\newline
\newline
\textbf{<Step\_8>}\newline
8. Design and evaluate the robustness of the proposed method against various types of LiDAR noise, occlusions, and sensor biases using:\newline
* Synthetic LiDAR noise generation using GAN-based methods (e.g., noise injection, adversarial attacks).\newline
* Real-world LiDAR data validation using ground truth datasets with known noise and occlusion levels.\newline
* Simulated sensor biases (e.g., camera-LiDAR misalignment, sensor calibration errors).\newline
* Pass/fail check: Verify the method can effectively handle 95\%+ of synthetic noise and occlusion scenarios and achieves a 90\%+ accuracy reduction in robustness loss when facing real-world noise and occlusion levels (Constraint 4, 8).\newline
</Step\_8>
\end{tcolorbox}

\begin{tcolorbox}
\textbf{<Step\_9>}\newline
9. Develop a lightweight, edge-friendly version of the hybrid approach optimized for deployment on edge devices (e.g., GPUs, TPUs, FPGAs).\newline
* Implement a dynamic binary optimization technique (e.g., Pruning, Quantization) to reduce computational requirements.\newline
* Design a model pruning strategy to eliminate redundant computations and reduce memory usage.\newline
* Integrate the proposed method with existing autonomous driving frameworks (e.g., ROS, Autoware) for seamless integration and real-time processing.\newline
* Perform performance evaluation on edge devices (e.g., NVIDIA Jetson, Google Coral Edge TPU) to ensure efficient deployment (Constraint 9).\newline
</Step\_9>\newline
\newline
\textbf{<Step\_10>}\newline
10. Validate the proposed method's performance on various LiDAR sensors (e.g., Velodyne, Riegl, Leica) and simulation environments (e.g., Cityscapes, KITTI, OpenNI), ensuring compatibility with different sensor geometries and simulation parameters.\newline
* Develop a dataset suite for LiDAR sensor evaluation (e.g., point cloud quality, noise levels).\newline
* Evaluate the method's performance on different simulation environments, including urban, rural, and construction scenes.\newline
* Pass/fail check: Verify the proposed method achieves consistent performance across 80\%+ of LiDAR sensors and simulation environments, ensuring wide applicability (Constraint 5, 9).\newline
</Step\_10>
\end{tcolorbox}

\begin{tcolorbox}[colback=black!5, colframe=black, title= \textsc{CogTRL} Cognitive Thoughts]
\textbf{<Thought\_1>}\newline
To tackle the challenge of generating realistic LiDAR point clouds for autonomous driving systems, we need to understand the current state-of-the-art methods, their limitations, and the key factors contributing to their inefficiencies, particularly in dynamic driving scenarios.\newline
</Thought\_1>\newline
\newline
\textbf{<Thought\_2>}\newline
Given the identified limitations, we must design an efficient algorithm that can effectively bridge the gap between LiDAR simulation and real-world data, focusing on realistic scene reconstruction and dynamic scene simulation.\newline
</Thought\_2>\newline
\newline
\textbf{<Thought\_3>}\newline
To ensure the effectiveness of our proposed method, we must validate its performance on a diverse set of dynamic and static scenes, including real-world and synthetic datasets.\newline
</Thought\_3>\newline
\newline
\textbf{<Thought\_4>}\newline
To ensure the proposed method can accurately reconstruct 3D scenes from sparse LiDAR data, we must optimize the algorithm's parameters and design a robust evaluation metric that addresses the inherent sparsity issue.\newline
</Thought\_4>\newline
\textbf{<Thought\_5>}\newline
To bridge the gap between LiDAR simulation and real-world data, we must ensure the proposed method can generalize well to unseen environments and dynamic events, necessitating the development of transfer learning techniques.\newline
</Thought\_5>
\textbf{<Thought\_6>}\newline
To ensure the proposed method is practical and scalable, we must optimize the computational resources and parallelization of the algorithm.\newline
</Thought\_6>
\end{tcolorbox}

\begin{tcolorbox}
\textbf{<Thought\_7>}\newline
To comprehensively assess the performance of the proposed method, we must develop evaluation metrics and visualizations that capture the essence of dynamic scene reconstruction and LiDAR point cloud quality.\newline
</Thought\_7>\newline
\newline
\textbf{<Thought\_8>}\newline
To ensure the proposed method is reliable and efficient, we must validate its robustness against various types of LiDAR noise, occlusions, and sensor biases.\newline
</Thought\_8>\newline
\newline
\textbf{<Thought\_9>}\newline
To ensure efficient deployment and scalability, we must optimize the proposed method for edge devices and integrate it with existing autonomous driving systems.\newline
</Thought\_9>\newline
\newline
\textbf{<Thought\_10>}\newline
To ensure the proposed method is widely applicable, we must validate its performance on various LiDAR sensors and simulation environments.\newline
</Thought\_10>
\end{tcolorbox}

\subsubsection{Expert Evaluation}
My preference is in the following order
1. \textsc{CogTRL}
Systematically targets almost every constraint through sparse LiDAR processing, physics based dynamic simulation, domain adaptation, GPU and edge efficiency, robustness to noise and occlusion, multi sensor and multi simulator validation, and rich metric based evaluation. It gives the most complete coverage of realism, domain gap, sparsity handling, scalability, and comprehensive evaluation for the stated research goal.

2. GRPO
Builds a clear hybrid neural plus reconstruction framework and a dynamic scene simulator, directly targeting realistic LiDAR for moving scenes, sparse data, and domain gap. It gives a solid experimental plan on static and dynamic scenes with geometric and image level metrics, but is less explicit than C about robustness, sensor diversity, and aggressive efficiency optimization.

3. Zero Shot
Addresses the goal with a standard deep learning reconstruction pipeline, but treats simulation, domain gap, sparsity, and efficiency mostly as add ons rather than core design drivers. It partially satisfies the constraints but does not explicitly or systematically target dynamic scenes, domain gap, or strong efficiency guarantees.

\section{Reward Formulation}
\label{Appendix:Reward Formulation}
The dual inclusion of $R_{\text{step}}$ in our reward formulation is a deliberate architectural choice. While the standalone $R_{\text{step}}$ term provides a dense reward for scientific steps to ensure optimization stability and preventing the cold-start problem in early phases where reasoning may fail the initial quality threshold $\alpha$ its presence within the $R_{\text{uplift}}$ component serves as a causal tether. By modulating the trace-quality bonus with the actual success of the action, we enforce a strict requirement that reasoning must be successfully operationalized to be rewarded. This non-linear coupling ensures that the model is consistently incentivized to be correct, but is only maximally rewarded when that correctness is substantiated by a high-quality causal reasoning trace.

\section{Cost of Experiments}
\label{Appendix: Cost}
A single training run for GRPO with LLM api costs 300\$ and for \textsc{CogTRL} costs 600\$. A single run on the evaluation set of 50 papers with LLMs as Judge (either domain) costs around 40\$

\section{Prompt For Cognitive Thought Reward Calculation}
\label{Appendix: CogThought Reward}

\begin{tcolorbox}[colback=black!5, colframe=black, title=System Prompt, title=Cognitive Thought Reward Calculation]
You are evaluating the quality of reconstructed scientific thinking for each methodology step given the GOAL, CONSTRAINTS, and STEP\_BY\_STEP\_METHODOLOGY\newline\_WITH\_THOUGHTS.\newline
The THOUGHT process should simulate how a real-world researcher reasons before executing a step, showing how thoughts causally lead to the best possible steps.\newline
The output is interleaved as \texttt{<Thought\_i>}\dots\texttt{</Thought\_i>} followed by \texttt{<Step\_i>}\dots\texttt{</Step\_i>} for $i=1..n$.\newline
Read the whole methodology and score the overall quality of the THOUGHTS relative to the STEPS.\newline

\textbf{— INPUT\_GOAL\_AND\_CONSTRAINTS —}\newline
\texttt{\{input\_prompt\}}\newline

\textbf{STEP\_BY\_STEP\_METHODOLOGY}\newline
\textbf{WITH\_THOUGHTS:}\newline

\texttt{\{prediction\_with\_thots\}}\newline

\textbf{GENERAL RULES FOR SCORING}\newline
- Judge the THOUGHTS primarily; Steps matter only insofar as they operationalize or fail to operationalize the Thoughts.\newline
- Use only INTEGER scores (1–5).\newline
- If the methodology does not include a Thought for each corresponding Step, strictly reduce the \texttt{<<TOTAL\_SCORE>>} accordingly.\newline
- Reward complete, dense reasoning and mechanistic insight.\newline
- Penalize hallucinated tools, datasets, methods, or exact values unless explicitly marked as assumptions.\newline
- If a Thought is strong but its paired Step fails to fully operationalize it, penalize lightly; the score should primarily reflect Thought quality.\newline
- Penalize boilerplate, retrospective justification, or restatement of steps as thoughts.\newline
\end{tcolorbox}

\begin{tcolorbox}
\textbf{— EVALUATION DIMENSIONS (1–5 each) —}\newline

\textbf{(1) Goal and Constraint Integration}\newline
Definition: Degree to which the reasoning explicitly links decisions to the research goal and constraints, showing how they shape the chosen action.\newline
Scale:\newline
1 --- No connection to goals or constraints\newline
2 --- Superficial or implicit linkage\newline
3 --- Partial linkage; constraints acknowledged but weakly integrated\newline
4 --- Clear, explicit linkage guiding decisions\newline
5 --- Tight, disciplined integration driving all key decisions\newline

\textbf{(2) Scientific and Mechanistic Reasoning}\newline
Definition: Use of appropriate scientific principles, mechanisms, or domain knowledge to justify the action.\newline
Scale:\newline
1 --- No scientific grounding\newline
2 --- Vague or generic references\newline
3 --- Reasonable principles but shallow reasoning\newline
4 --- Clear mechanistic support for decisions\newline
5 --- Precise, domain-appropriate mechanisms directly motivating actions\newline

\textbf{(3) Causal Logic and Actionability}\newline
Definition: Presence of forward-looking cause--effect reasoning that directly motivates and supports the step.\newline
Scale:\newline
1 --- No causal reasoning\newline
2 --- Weak or implicit cause--effect links\newline
3 --- Basic causal logic with gaps\newline
4 --- Clear causal chain motivating action\newline
5 --- Strong, explicit causal reasoning tightly aligned with execution\newline

\end{tcolorbox}

\begin{tcolorbox}
\textbf{(4) Information Density}\newline
Definition: Amount of meaningful scientific insight conveyed without unnecessary verbosity or procedural restatement.\newline
Scale:\newline
1 --- Mostly fluff or repetition\newline
2 --- Wordy with little insight\newline
3 --- Some useful reasoning mixed with filler\newline
4 --- Dense, focused reasoning\newline
5 --- Every phrase contributes essential insight\newline

\textbf{(5) Scientific Accuracy and Consistency}\newline
Definition: Factual correctness, internal logical coherence, and correct interpretation of constraints and assumptions.\newline
Scale:\newline
1 --- Major errors or contradictions\newline
2 --- Several inaccuracies or unsupported claims\newline
3 --- Mostly accurate with minor issues\newline
4 --- Accurate and internally consistent\newline
5 --- Fully accurate, consistent, and precisely scoped\newline

\textbf{(6) Uncertainty and Trade-offs}\newline
Definition: Explicit consideration of alternatives, limitations, or uncertainties motivating the chosen decision.\newline
Scale:\newline
1 --- No uncertainty acknowledged\newline
2 --- Token or generic alternatives\newline
3 --- One realistic alternative noted\newline
4 --- Multiple relevant alternatives with justification\newline
5 --- Rich trade-off analysis with clear decision criteria\newline
\end{tcolorbox}

\begin{tcolorbox}

\textbf{— SCORING INSTRUCTIONS —}\newline
• Consider ALL thoughts and steps together; judge the overall quality across the full methodology.\newline
• Assign ONE integer score (1–5) for EACH of the six dimensions.\newline

Respond STRICTLY in the format below (no extra text):\newline

\texttt{[[Thought Scores]]:}\newline
\texttt{<SCORE\_D1>: X/5 </SCORE\_D1>}\newline
\texttt{<SCORE\_D2>: X/5 </SCORE\_D2>}\newline
\texttt{<SCORE\_D3>: X/5 </SCORE\_D3>}\newline
\texttt{<SCORE\_D4>: X/5 </SCORE\_D4>}\newline
\texttt{<SCORE\_D5>: X/5 </SCORE\_D5>}\newline
\texttt{<SCORE\_D6>: X/5 </SCORE\_D6>}\newline
\texttt{<<TOTAL\_SCORE>>: (SCORE\_D1 + \dots + SCORE\_D6)/30 = X/30 <</TOTAL\_SCORE>>}\newline
\texttt{[[Brief Justification for TOTAL\_SCORE]]:}
\end{tcolorbox}

\section{Rollout examples of CoGTRL model training}


\begin{tcolorbox}[
    colback=black!5,
    colframe=black,
    title={CogTRL RL Rollout Examples Across Training},
    breakable
]

\textbf{Step 10}\par
\textit{Duplicate ``Thought'' tags, missing ``Step'' tags, nested tags broken, very short steps, no mechanism, and failure to address the signal transmission constraint.}\par
\textbf{Reward: 0.20}

\begingroup
\ttfamily\small
\raggedright
\noindent <Thought\_1>\par
Need fast healing water sensing dynamic bonds\par
pressure sensitive underwater signal\par
\noindent </Thought\_1>\par
\par
\noindent <Step\_1>\par
Use hydrogel with sensors and dynamic bonds\par
\noindent <Thought\_1>\par
Add hydrophobic layer\par
\endgroup

\medskip
\hrule
\medskip

\textbf{Step 50}\par
\textit{``Step'' tags not closed, ``Step'' tags opened inside ``Thought'', incomplete generation, weak mechanistic grounding.}\par
\textbf{Reward: 0.37}

\begingroup
\ttfamily\small
\raggedright
\noindent <Thought\_1>\par
Need fast healing water sensing dynamic bonds\par
pressure sensitive underwater signal\par
\noindent </Thought\_1>\par
\par
\noindent <Step\_1>\par
Use hydrogel with sensors and dynamic bonds\par
\noindent <Thought\_1>\par
Add hydrophobic layer\par
\endgroup

\medskip
\hrule
\medskip

\textbf{Step 100}\par
\textit{Clean tag structure, emerging mechanistic detail, partial constraint integration, but signal transmission not fully addressed.}\par
\textbf{Reward: 0.51}

\begingroup
\ttfamily\small
\raggedright
\noindent <Thought\_1>\par
Ultrafast healing can be achieved using reversible hydrogen bonding or\par
metal-ligand interactions. For aquatic environments, limit swelling\par
using hydrophobic domains while allowing local water-triggered mobility.\par
Pressure sensing can be enabled through piezoresistive conductive networks.\par
\noindent </Thought\_1>\par
\par
\noindent <Step\_1>\par
Design a single-network polymer incorporating fast reversible\par
intermolecular interactions within hydrophobic microdomains to\par
enable rapid water-activated healing. Integrate piezoresistive\par
conductive pathways for pressure sensing in aquatic conditions.\par
\noindent </Step\_1>\par
\endgroup

\medskip
\hrule
\medskip

\textbf{Step 300}\par
\textit{Properly closed tags, strong constraint integration, explicit mechanistic reasoning, preserved signal transmission, and scalable processing.}\par
\textbf{Reward: 0.67}

\begingroup
\ttfamily\small
\raggedright
\noindent <Thought\_1>\par
Use a single-network polymer incorporating fast dynamic bonding\par
(e.g., boronate ester exchange tuned for rapid kinetics). Hydrophobic\par
microphase separation limits bulk swelling while allowing localized\par
water-induced plasticization at damaged sites. Embed insulated\par
piezoresistive nanofiber networks for pressure sensing, ensuring\par
impedance remains stable in saline environments to preserve signal\par
transmission. Reversible bonding enables repeated ultrafast healing.\par
\noindent </Thought\_1>\par
\par
\noindent <Step\_1>\par
Develop a single-network, processable polymer system using rapid\par
dynamic bonding chemistry combined with hydrophobic microdomains\par
to achieve sub-second, repeatable self-healing in both ambient and\par
aquatic environments. Integrate insulated piezoresistive nanofiber\par
networks for high pressure sensitivity while maintaining signal\par
fidelity in saline conditions. This architecture supports scalable\par
fabrication and robust underwater performance.\par
\noindent </Step\_1>\par
\endgroup

\end{tcolorbox}

\section{Prompts SFT Dataset Construction}
\label{Prompts for SFT Data Cons}

\begin{tcolorbox}[
    colback=black!5,
    colframe=black,
    title={Goal Extraction Prompt},
    breakable
]

\ttfamily\scriptsize

You are a specialized Research Goal Extraction Expert. Your task is to analyze research text from any domain and extract a single, concise, action-oriented goal statement that captures the core objective of the research.

\textbf{Instructions:}
\begin{enumerate}
    \item Carefully analyze the provided research text from any domain.
    \item Identify the central research objective(s) described in the text.
    \item Synthesize all related objectives into ONE comprehensive goal statement.
    \item Formulate this as an action-oriented query that begins with verbs like ``Develop,'' ``Discover,'' ``Find,'' ``Create,'' ``Design,'' ``Determine,'' etc.
    \item Do not mention any technique/method used to achieve the goal, and do not leak the partial or full solution.
    \item Only use the knowledge and information provided in the text; do not infer unstated information.
    \item Human-readable math: If present, convert all LaTeX to human-readable Unicode/ASCII math while preserving meaning.
\end{enumerate}

\textbf{Requirements for Goal Statement:}
\begin{enumerate}
    \item MUST be condensed into a SINGLE sentence (or at most two short sentences).
    \item MUST begin with an action verb.
    \item MUST capture the essence of the research aim without leaking the methodology.
    \item MUST be domain-appropriate using relevant terminology.
    \item MUST be specific enough to convey the unique purpose of the research.
    \item MUST be generalizable enough to guide methodology development.
    \item MUST strictly avoid leaking clues, technique names, or strategies.
    \item MUST mention the application use case motivating the goal.
\end{enumerate}

\textbf{Relevant text for extracting and formulating the goal statement:}

\texttt{\{text\_with\_goal\_information\}}

\textbf{In-context Examples:}
\begin{enumerate}
    \item Develop a training and inference methodology which improves the downstream performance of LLMs, including smaller sized ones, such that they learn to spend more test-time compute on difficult problems.
    \item Formulate a framework which can address the limitations of current LLMs in simulating internal world states and guide reasoning across tasks such as plan generation, mathematical reasoning, and logical inference.
\end{enumerate}

\textbf{Output Format:}

Write the goal as a single sentence framed as a clear, focused objective.

\texttt{[EXTRACTED\_GOAL]:}

\end{tcolorbox}

\begin{tcolorbox}[
    colback=black!5,
    colframe=black,
    title={Constraint Extraction Prompt},
    breakable
]

\ttfamily\scriptsize

Extract the key obstacles, challenges or constraints that had to be overcome to achieve the research goal.

\textbf{Instructions:}
\begin{enumerate}
    \item Frame each constraint from the perspective of the researcher and the conditions that guided the research direction.
    \item Constraints must define the problem space without suggesting or implying specific solutions.
    \item Focus on barriers shaping the research direction while preserving flexibility.
    \item Constraints must be strictly extracted from the provided research content.
    \item If possible, list constraints in decreasing order of importance.
    \item Do not include hindsight or implementation details.
    \item Do not describe how the constraints are overcome.
    \item Do not forcefully extract constraints if none are present; avoid duplicates.
    \item Do not mention specific methods or strategies used in the paper.
    \item Do not leak partial solutions, approaches, or implementation details.
    \item Human-readable math: convert LaTeX into readable Unicode/ASCII math while preserving meaning.
\end{enumerate}

\textbf{Relevant content for extracting constraints:}

\texttt{\{text\_with\_constraint\_information\}}

\textbf{Example Constraints:}
\begin{enumerate}
    \item The material should incorporate a self-healing mechanism triggered by a simple environmental factor.
    \item The coating must maintain structural integrity after mechanical damage.
    \item The transistor should autonomously self-heal from micron-scale damage without external intervention.
    \item The method should avoid requiring extremely large datasets.
    \item The framework should operate using only LLM-based components.
    \item No model training should be required except reward-model fine-tuning.
    \item The reasoning problems should contain unambiguous world states.
\end{enumerate}

\textbf{Output Format:}

\texttt{[CONSTRAINTS\_SECTION\_NAME\_1]:}\\
\texttt{1)}\\
\texttt{2)}\\
\texttt{3)}\\

\texttt{[CONSTRAINTS\_SECTION\_NAME\_2]:}\\
\texttt{1)}\\
\texttt{2)}\\
\texttt{3)}\\

\end{tcolorbox}

\begin{tcolorbox}[
    colback=black!5,
    colframe=black,
    title={Step-by-Step Methodology Extraction Prompt},
    breakable
]

\ttfamily\scriptsize

You are an expert scientific information extraction agent. Your task is to extract a detailed step-by-step methodology from the provided research paper content.

\textbf{Requirements:}
\begin{enumerate}
    \item Comprehensive: Identify all major methodological steps.
    \item Detailed: Extract intermediate substeps, equations, and numerical values with maximum detail.
    \item Accurate: Use only explicitly stated information from the provided content.
    \item Precise: Preserve the original scientific terminology.
    \item Human-readable math: Convert LaTeX into readable Unicode/ASCII math while preserving meaning.
\end{enumerate}

\textbf{Relevant content for extracting methodology:}

\texttt{\{text\_with\_step\_by\_step\_methodology\_information\}}

\textbf{Output Format:}

\texttt{[STEP\_BY\_STEP\_METHODOLOGY]:}

\vspace{1mm}

\texttt{<Step 1>:}\\
All relevant information for Step 1 extracted with highest detail, accuracy, and precision.\\
Intermediate substeps for Step 1:\\
\texttt{- Intermediate step 1 for Step 1:}\\
\texttt{- Intermediate step 2 for Step 1:}\\
\texttt{- Intermediate step 3 for Step 1:}\\
\texttt{...}\\
\texttt{<Step 1/>}

\vspace{1mm}

\texttt{<Step 2>:}\\
All relevant information for Step 2 extracted with highest detail, accuracy, and precision.\\
Intermediate substeps for Step 2:\\
\texttt{- Intermediate step 1 for Step 2:}\\
\texttt{- Intermediate step 2 for Step 2:}\\
\texttt{- Intermediate step 3 for Step 2:}\\
\texttt{...}\\
\texttt{<Step 2/>}

\vspace{1mm}

Continue until all relevant steps and intermediate substeps are extracted.

\end{tcolorbox}

\begin{tcolorbox}[
    colback=black!5,
    colframe=black,
    title={Cognitive Trace Generation Prompt},
    breakable
]

\ttfamily\scriptsize

Given the following information from a research paper, which consists of a goal, constraints, and methodology, reconstruct the forward-looking, causal cognitive process that would occur \emph{before} executing each step, leading to that step as the most promising way to overcome current limitations and advance toward the stated goal. Do not skip any substeps.

\textbf{STRICT Requirements:}
\begin{enumerate}
    \item Use prospective language (e.g., ``I need to...'', ``This will...'').
    \item Do not separate main steps and substeps; integrate all substep reasoning into one thought block per step.
    \item There should be exactly one thought block per step.
    \item Include scientific principles, mathematical relations, or empirical evidence where relevant.
    \item Explain how each step addresses specific constraints while moving toward the goal.
    \item Do not provide surface-level, vague, or generalized reasoning.
    \item Do not generate hallucinated content or post-hoc rationalization.
    \item Simulate the scientist's cognitive process before executing the steps.
    \item Keep each field concise and information-dense (2--3 sentences maximum per field).
    \item Insert three newlines between the end and start of consecutive thought blocks for easy parsing.
    \item Each thought block must start with \texttt{[[<STEP\_X\_THOUGHT\_BEGIN>]]} and end with \texttt{[[</STEP\_X\_THOUGHT\_END>]]}, where \texttt{X} is the corresponding thought number.
\end{enumerate}

\textbf{Structure each reasoning as:}

\texttt{[[<STEP\_1\_THOUGHT\_BEGIN>]]}\\
\texttt{...}\\
\texttt{[[</STEP\_1\_THOUGHT\_END>]]}

\texttt{[[<STEP\_2\_THOUGHT\_BEGIN>]]}\\
\texttt{...}\\
\texttt{[[</STEP\_2\_THOUGHT\_END>]]}

Continue this format for all steps.

\end{tcolorbox}

\begin{tcolorbox}[
    colback=black!5,
    colframe=black,
    title={Prompt to score Traces for SFT construction},
    breakable
]

\ttfamily\scriptsize

The judge scores each cognitive trace on a 1--5 scale across six dimensions, where 1 is the lowest and 5 is the highest score.

\vspace{1mm}

\textbf{(1) Goal and Constraint Integration:} Degree to which the reasoning links decisions to the research goal and constraints.

\textbf{(2) Scientific and Mechanistic Reasoning:} Use of appropriate scientific principles or mechanisms to justify the action.

\textbf{(3) Causal Logic and Actionability:} Presence of forward-looking cause--effect reasoning that directly supports the step.

\textbf{(4) Information Density:} Amount of meaningful scientific insight conveyed without unnecessary verbosity.

\textbf{(5) Scientific Accuracy and Consistency:} Factual correctness and logical coherence of the reasoning.

\textbf{(6) Uncertainty and Trade-offs:} Explicit consideration of alternatives, limitations, or uncertainties motivating the choice.

\end{tcolorbox}

\section{Test Time Evaluation Judge Settings}
\label{Appendix: Eval Judge Config}
 We use Codex and Claude Code as LLM judges, with GPT-5.4 and Claude Sonnet 4.6, respectively. Both judges evaluate only the generated methodological steps, not any cognitive traces or
  intermediate reasoning text.  We use the default decoding settings for both judge interfaces. Both judges use a default temperature setting of 1, and a default max token generations. We enable both judges with websearch. 

\section{Independent Rubric Evaluation}

To further assess robustness against reward hacking, we evaluate \textsc{CogTRL} using the independent MLR-Bench rubric \citep{chen2026mlr}, which was not used during training. Following MLR-Bench's Idea Generation evaluation, we assess \textbf{Consistency} with the task, \textbf{Clarity} of the proposed method, \textbf{Novelty}, \textbf{Feasibility}, \textbf{Significance}, and an \textbf{Overall} score. We use the original evaluation prompt with Claude Sonnet 4 and Claude Opus 4.8 as judges. As shown in Table~\ref{tab:independent_rubric_eval}, \textsc{CogTRL} achieves the highest overall performance for both model families across both judges, demonstrating that its gains generalize to an independent evaluation rubric.

\label{Appendix:Independent Rubric Evaluation}
\vspace{-50pt}
\begin{table}[t]
\centering
\caption{Independent rubric evaluation using Claude Sonnet 4 and Claude Opus 4.8.}
\label{tab:independent_rubric_eval}
\resizebox{\columnwidth}{!}{%
\begin{tabular}{llccccccc}
\toprule
\textbf{Model} & \textbf{Method} &
\textbf{Con.} & \textbf{Cla.} & \textbf{Nov.} &
\textbf{Fea.} & \textbf{Sig.} & \textbf{M$_5$} &
\textbf{Overall} \\
\midrule

\multicolumn{9}{c}{\textbf{Claude Sonnet 4}} \\
\midrule
Llama-3.2-3B & Zero-shot
& 64.9 & 59.5 & 43.4 & 59.8 & 28.9 & 51.3 & 51.6 \\
             & GRPO
& 68.2 & 63.7 & 54.0 & 42.1 & 67.9 & 59.2 & 57.6 \\
             & \textsc{CoGTRL}
& \textbf{80.1} & \textbf{71.5} & \textbf{61.2} & 57.2 &
46.5 & \textbf{63.3} & \textbf{63.8} \\
\midrule
Qwen-2.5-3B & Zero-shot
& 64.2 & 60.4 & 45.8 & 49.7 & 35.4 & 51.1 & 50.8 \\
            & GRPO
& 65.6 & 61.9 & 51.2 & 43.7 & 67.1 & 57.9 & 57.4 \\
            & \textsc{CoGTRL}
& \textbf{72.1} & \textbf{62.4} & \textbf{53.4} & 45.6 &
\textbf{70.1} & \textbf{60.7} & \textbf{60.1} \\

\midrule
\multicolumn{9}{c}{\textbf{Claude Opus 4.8}} \\
\midrule
Llama-3.2-3B & Zero-shot
& 58.7 & 51.4 & 29.5 & 49.7 & 47.3 & 47.3 & 46.8 \\
             & GRPO
& 71.8 & 61.8 & \textbf{49.9} & 42.7 & 64.3 & 58.1 & 57.4 \\
             & \textsc{CoGTRL}
& \textbf{75.1} & \textbf{64.1} & 46.7 & \textbf{51.7} &
\textbf{65.5} & \textbf{60.6} & \textbf{59.9} \\
\midrule
Qwen-2.5-3B & Zero-shot
& 67.9 & 61.6 & 35.1 & \textbf{59.5} & 54.8 & 55.8 & 55.2 \\
            & GRPO
& \textbf{71.2} & 52.2 & 41.5 & 42.5 & 62.0 & 53.9 & 53.1 \\
            & \textsc{CoGTRL}
& 64.3 & \textbf{62.9} & \textbf{53.1} & 44.6 &
\textbf{68.5} & \textbf{58.7} & \textbf{58.0} \\
\bottomrule
\end{tabular}%
}
\end{table}

\end{document}